\documentclass[lettersize,journal]{IEEEtran}
\usepackage{amsmath,amsfonts}
\usepackage{algorithmic}
\usepackage{algorithm}
\usepackage{array}
\usepackage[caption=false,font=normalsize,labelfont=sf,textfont=sf]{subfig}
\usepackage{textcomp}
\usepackage{stfloats}
\usepackage{url}
\usepackage{verbatim}
\usepackage{graphicx}
\usepackage{cite}
\usepackage{amsthm}
\usepackage{amssymb}
\usepackage{xcolor}
\usepackage{multirow}
\usepackage{booktabs}

\usepackage{algorithm}
\usepackage{algorithmic}

\newtheorem{theorem}{Theorem}
\newtheorem*{theorem*}{Theorem}
\newtheorem{proposition}{Proposition}

\begin{document}

\title{A Unified and Fast-Sampling Diffusion Bridge Framework via Stochastic Optimal Control}

\author{Mokai Pan, Kaizhen Zhu, Yuexin Ma, Yanwei Fu, Jingyi Yu, Jingya Wang, Ye Shi%, ~\IEEEmembership{Member,~IEEE,}
        % <-this % stops a space
% \thanks{This work was supported by National Natural Science Foundation of China (62303319, 62406195), Shanghai Local College Capacity Building Program (23010503100), ShanghaiTech AI4S Initiative SHTAI4S202404, HPC Platform of ShanghaiTech University, Core Facility Platform of Computer Science and Communication of ShanghaiTech University, and MoE Key Laboratory of Intelligent Perception and Human-Machine Collaboration (ShanghaiTech University), and Shanghai Engineering Research Center of Intelligent Vision and Imaging.}% <-this % stops a space
\thanks{Mokai Pan and Kaizhen Zhu contributed equally to this work. Corresponding author: Ye Shi.}
\thanks{Mokai Pan, Kaizhen Zhu, Yuexin Ma, Jingyi Yu, Jingya Wang, and Ye Shi are with the School of Information Science and Technology, ShanghaiTech University, Shanghai 201210, China (e-mail: \{panmk2025, zhukzh2024, mayuexin, yujingyi, wangjingya, shiye\}@shanghaitech.edu.cn)}
\thanks{Yanwei Fu is with the School of Data Science, Fudan University, Shanghai 200433, China (e-mail: yanweifu@fudan.edu.cn).}}

% \markboth{IEEE TRANSACTIONS ON PATTERN ANALYSIS AND MACHINE INTELLIGENCE}


\maketitle

\begin{abstract}
Recent advances in diffusion bridge models leverage Doob’s $h$-transform to establish fixed endpoints between distributions, demonstrating promising results in image translation and restoration tasks. However, these approaches often produce blurred or excessively smoothed image details and lack a comprehensive theoretical foundation to explain these shortcomings. To address these limitations, we propose UniDB, a unified and fast-sampling framework for diffusion bridges based on Stochastic Optimal Control (SOC). We reformulate the problem through an SOC-based optimization, proving that existing diffusion bridges employing Doob’s $h$-transform constitute a special case, emerging when the terminal penalty coefficient in the SOC cost function tends to infinity. By incorporating a tunable terminal penalty coefficient, UniDB achieves an optimal balance between control costs and terminal penalties, substantially improving detail preservation and output quality. To avoid computationally expensive costs of iterative Euler sampling methods in UniDB, we design a training-free accelerated algorithm by deriving exact closed-form solutions for UniDB’s reverse-time SDE. It is further complemented by replacing conventional noise prediction with a more stable data prediction model, along with an SDE-Corrector mechanism that maintains perceptual quality for low-step regimes, effectively reducing error accumulation. Extensive experiments across diverse image restoration tasks validate the superiority and adaptability of the proposed framework, bridging the gap between theoretical generality and practical efficiency. Our code is available online\footnote{https://github.com/2769433owo/UniDB-plusplus}.
\end{abstract}


\section{Introduction}

\begin{figure*}[t]
    \centering
    \includegraphics[width=0.95\textwidth]{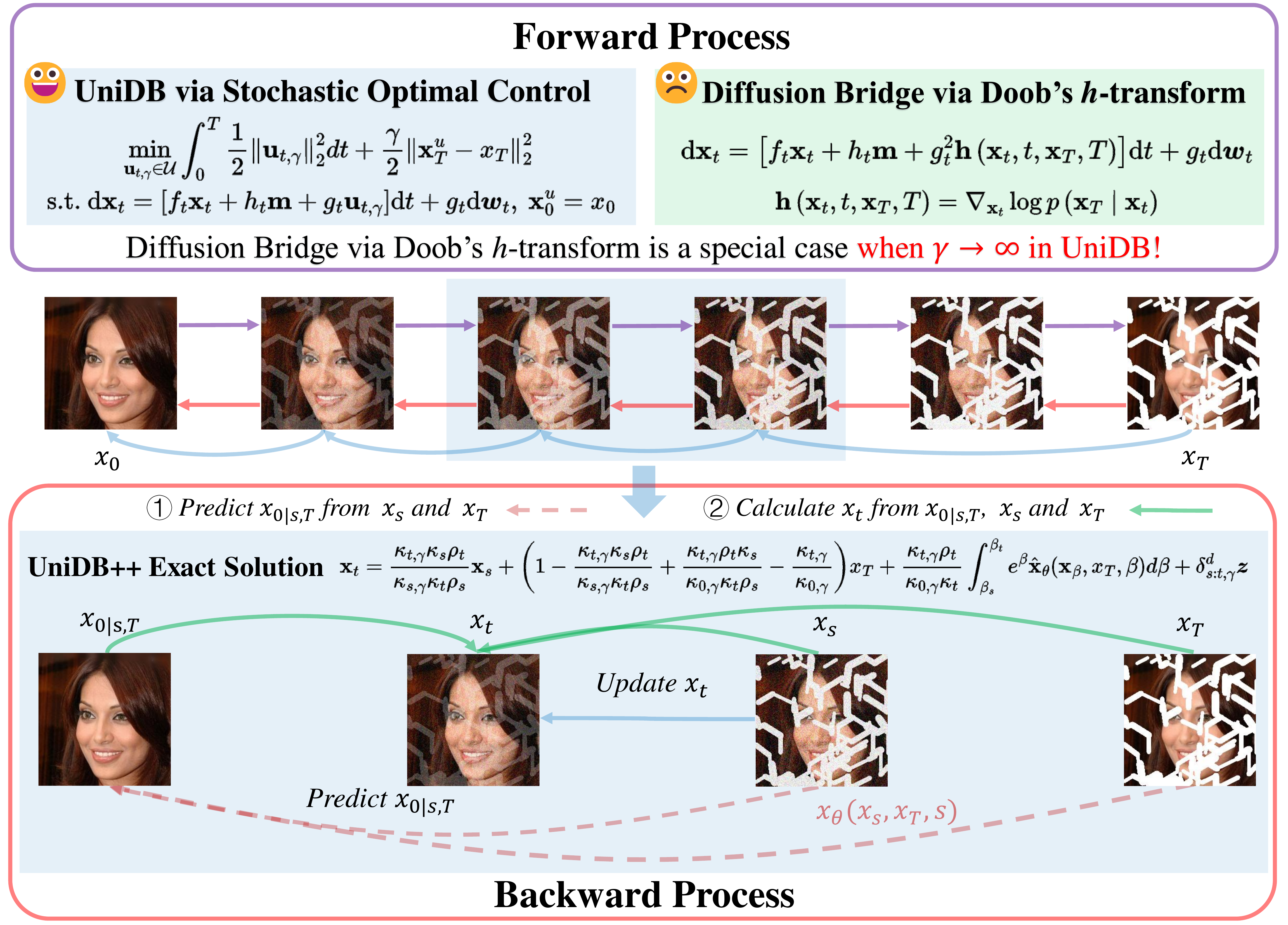}
    \vspace{-3mm}
    \caption{We formulate the forward process of diffusion bridges as a stochastic optimal control problem and employs the penalty coefficient $\gamma$, balancing realistic SDE trajectories and target endpoint matching, to produce images with more realistic details. We also find that Doob’s $h$-transform is a special case of UniDB when $\gamma \rightarrow \infty$. To overcome the low-efficiency in sampling process, we develop a fast training-free sampling method UniDB++ by deriving the exact solution of the reverse SDE with data prediction model, which directly estimates the fixed and smooth target $\textbf{x}_0$, ensuring stable predictions of the solvers even under few sampling-step regimes.}
    \label{fig_main}
\end{figure*}

\IEEEPARstart{D}{iffusion} models have been widely used in a variety of applications, demonstrating remarkable capabilities and promising results in numerous tasks such as image generation \cite{DDPM, score-based, diffir}, inverse problem solving \cite{DPS, DSG, DDRM}, video generation \cite{vd1, vd2}, imitation learning \cite{afforddp, dp, dp3, goal}, and reinforcement learning \cite{QVPO, rl1, rl2, rl3, rl4}, etc. However, since the prior distribution of standard diffusion models is assumed to be standard Gaussian noise, they are difficult to achieve the conversion between any two distributions, particularly exhibiting inherent limitations in various image translation, editing, and restoration tasks.

To overcome this problem, conditional diffusion models \cite{classifier, classifier-free, gibbsddrm, CCDM, DPS, HM, DSG, RB} focus on meticulously designing conditioning mechanisms during training process or introducing training-free classifier or loss guidance tricks to facilitate conditional sampling, ensuring output alignment with a target distribution. However, these methods can be cumbersome, expensive, and may introduce manifold deviations during the sampling process or fitting errors from neural networks. 

To avoid the limitations, some approaches have made attempts to learn the straight mapping between two arbitrary distributions. On one hand, Diffusion Schrödinger Bridge \cite{DSB2, DSB1, ADSB}  seeks an optimal stochastic process for inter-distribution transport under KL divergence minimization, yet its computational complexity from iterative procedures restricts practical applicability. On the other hand, DDBMs \cite{DDBMs} mathematically combine Doob’s $h$-transform \cite{ASDE} with the simple linear stochastic differential equation (SDE), Variance Exploding (VE) and Variance Preserving (VP) SDE \cite{score-based}, successfully solving the distribution translation tasks. However, the forward SDE in DDBMs lacks the mean information of the terminal distribution, which restricts the quality of the generated images. GOUB \cite{GOUB} further extends DDBMs by integrating Mean-Reverting SDE (MR-SDE) with Doob's $h$-transform, achieving better image restoration results. Despite the empirical success of Doob's $h$-transform in building diffusion bridges, its theoretical underpinnings remain obscure, and its practical application often sacrifices critical high-frequency details—such as sharp edges and textures—resulting in outputs with perceptually degraded, oversmoothed artifacts.

To address these limitations, we revisit diffusion bridges through an SOC-based optimization. We demonstrate that Doob’s $h$-transform is a special case of SOC theory under the condition that the terminal penalty coefficient approaches infinity, thereby unifying and generalizing existing diffusion bridges. The utilized penalty coefficient successfully enhanced the authenticity of the generated outputs. However, the adopted naive Euler method for sampling usually demands hundreds of sampling steps for image restoration tasks, resulting in suboptimal efficiency. This inefficiency poses a substantial barrier to the practical deployment in real-time applications. To bridge this gap, we further develop a fast training-free sampling method. We replace the conventional noise prediction model with the more stable data prediction model which directly estimates the fixed and smooth target $\textbf{x}_0$, ensuring stable predictions of the solvers even under few sampling-step regimes. By deriving the exact closed-form solutions for the reverse-time SDE along with an SDE-Corrector mechanism that improves perceptual quality for low-step regimes (5-10 steps), the error accumulation inherent in Euler approximations is effectively reduced. The new acceleration algorithm demonstrates about 5$\times$ and up to 20$\times$ faster sampling and state-of-the-art performance in various image restoration tasks. Our main contributions are as follows:

\begin{itemize}

\item We introduce UniDB, a novel unified and fast-sampling diffusion bridge framework based on stochastic optimal control, which offers a comprehensive understanding and extension of Doob’s $h$-transform by incorporating general forward SDE forms. We derive closed-form solutions for the SOC problem, demonstrating that Doob’s $h$-transform is merely a special case when the terminal penalty coefficient in SOC approaches infinity. This insight reveals inherent limitations in the existing diffusion bridge approaches, which our method overcomes.

\item To improve the inefficient inference, we propose UniDB++, a fast training-free sampling algorithm. Our key advancement comes from deriving exact closed-form solutions for the reverse-time SDE, effectively reducing the error accumulation inherent in Euler approximations. We replace the conventional noise prediction with the more stable data prediction model, along with an SDE-Corrector mechanism that improves perceptual quality for extremely low-step (5$\sim$8 steps) regimes without exceptional costs. 

\item Our framework achieves state-of-the-art performance in various image restoration tasks, including super-resolution, deraining, inpainting, and raindrop restoration. UniDB++ shows 5$\times$ and up to 20$\times$ faster sampling compared to UniDB while maintaining high generation quality. This demonstrates the superiority of our methods and highlighting the model’s exceptional image quality and its versatility across a wide range of scenarios. 
\end{itemize}

% Meanwhile, existing acceleration techniques for diffusion bridges such as DBIMs \cite{DBIM} and CDBMs \cite{CDBMs}, which are fundamentally rooted in DDBMs and Doob's $h$-transform, cannot be directly applied to enhance the sampling efficiency of more generalized bridge frameworks like GOUB and UniDB. This limitation stems from two fundamental distinctions: 1) both GOUB and UniDB explicitly incorporate the mean information of the terminal distribution into their SDE formulations; 2) UniDB further introduces the optimal controller with penalty coefficient different from the $h$ function in Doob's $h$-transform, both of which result in completely different forms of SDEs. 

A preliminary version of our study has appeared as UniDB \cite{UniDB}, as a \textbf{spotlight} presentation in ICML 2025. Compared to UniDB, we extend our previous work in the following aspects: (1) We derive the closed-form solutions for UniDB’s reverse-time SDE, complemented by replacing conventional noise prediction with the more stable data prediction model, along with an SDE-Corrector mechanism that improves perceptual quality for low-step regimes. The exact solution allows the linear component to be computed analytically, leaving only integral with neural network term subject to approximation. (2) We systematically extend the experimental validation by evaluating UniDB on the Raindrop Restoration task and UniDB++, covering all tasks the same as UniDB. (3) We additionally perform rigorous ablation studies on UniDB++ to assess the impact of its different orders and the SDE-Corrector mechanism. Comprehensive experimental evaluations confirm that UniDB++ achieves superior and state-of-the-art results in both sample efficiency and image quality in various image restoration tasks.

\section{Related Work}

\begin{figure*}[t] 
    \centering
    \includegraphics[width=0.99\textwidth]{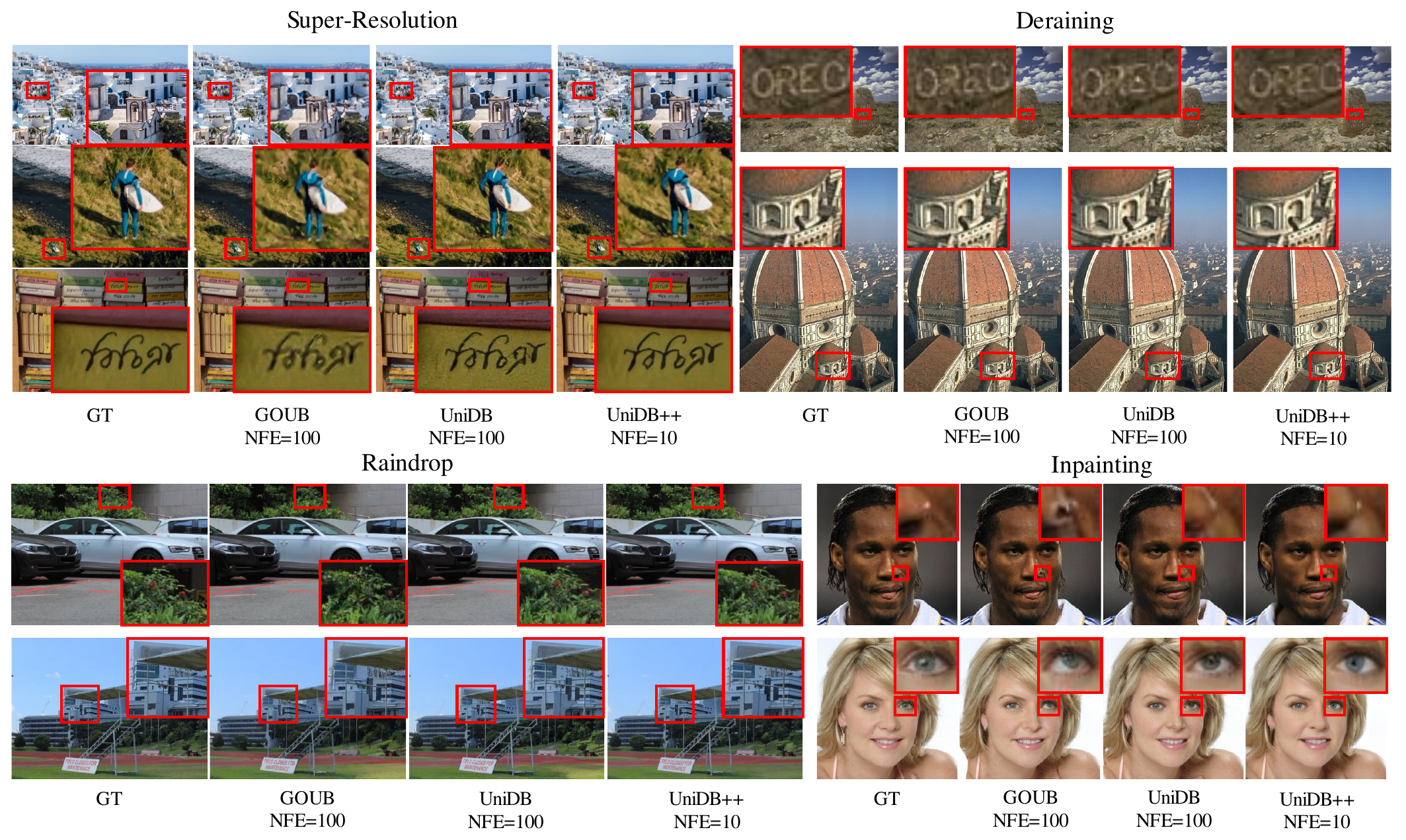}
    \vspace{-4mm}
    \caption{Qualitative comparison of visual results among Ground-Truth (GT) images, GOUB, UniDB (ours) and UniDB++ (ours) on Image 4$\times$Super-Resolution, Image Deraining, Image RainDrop Restoration, and Image Inpainting tasks with zoomed-in image local regions.}
    \label{fig_main_experiment}
\end{figure*}

\subsection{Diffusion Bridges}
Diffusion bridge models construct a stochastic process to learn the straight transformation between two arbitrary distributions. One line of the research is Diffusion Schrödinger Bridge \cite{I2SB, DSB1, DSB2, chen2021likelihood, ADSB}, which aims to determine a stochastic process that facilitates probabilistic transport between given initial and terminal distribution while minimizing the entropy cost. However, the training process is often complex, resulting in slow convergence and limited model fitting capabilities. The other line of the research is Diffusion Bridge with Doob's $h$-transform. Notably, DDBMs \cite{DDBMs} pioneered by adopting a linear SDE together with Doob's $h$-transform to build direct diffusion bridges. GOUB \cite{GOUB} further extended this framework by incorporating Doob's $h$-transform into a mean-reverting SDE, attaining state-of-the-art results in image restoration tasks. Despite these empirical successes, the theoretical foundations of Doob's $h$-transform remain insufficiently explored. Moreover, these methods often suffer from blurred or oversmoothed features in generated images, particularly pertaining to the problem of capturing high-frequency details and perceptual fidelity. 

\subsection{Diffusion with Stochastic Optimal Control}
The integration of SOC principles into diffusion models has emerged as a promising paradigm for guiding distribution transitions. Prior work includes DIS \cite{OCdiffusion}, which first constructed a theoretical connection between SOC and diffusion processes, and RB-Modulation \cite{RB}, which implemented SOC via a simplified SDE to enable training-free style transfer using pre-trained models. Closely related to our work, DBFS \cite{DBFS} leveraged SOC to construct diffusion bridges in infinite-dimensional function spaces and also established equivalence between SOC and Doob's $h$-transform. However, DBFS primarily leverage SOC to extend Doob's $h$-transform to infinite-dimensional function spaces, without addressing its intrinsic limitations, while UniDB reveals a critical insight that Doob's $h$-transform corresponds to a suboptimal solution that can inherently lead to artifacts such as blurred or distorted details.

\subsection{Diffusion Solvers for Fast Sampling}
Methods for accelerating the sampling of diffusion models fall primarily into training-free and training-based algorithms. Training-based diffusion acceleration methods often employ distillation techniques such as \cite{Adistill, ConsistencyM, S4S}. Training-free methods like DDIMs \cite{DDIM}, DPM-Solver \cite{dpm-solver}, DPM-Solver++ \cite{dpm-solver-plus}, DPM-Solver-v3 \cite{dpm-solver-v3}, and UniPC \cite{Unipc} constitute a family of fast numerical SDE or ODE solvers specifically designed for standard diffusion models. Unlike them, MaRS \cite{MaRS} proposed a special numerical solver to accelerate the sampling of reverse-time MR-SDE and achieves promising speedup in various image restoration tasks. However, these acceleration methods are all designed for diffusion models, not diffusion bridges at all. Specifically for diffusion bridges, CDBMs \cite{CDBMs} improve the efficiency of DDBMs through consistency functions including Consistency Bridge Distillation (CBD) and Consistency Bridge Training (CBT), while BBDM \cite{BBDM}, DBIMs \cite{DBIM} and I3SB \cite{wang2025implicit} construct a series of non-Markovian diffusion bridges based on Brownian Bridge, DDBMs, and I2SB, respectively. They all achieved faster and better sampling results. However, these bridges lack both the mean information of the terminal distribution in their SDE formulation and the penalty coefficient derived from SOC theory, both of which result in completely different SDE forms, making their acceleration methods fundamentally incompatible with UniDB’s sampling procedures.

\section{Preliminaries}
\subsection{Denoising Diffusion Bridge Models}
Diffusion models \cite{DDPM, score-based} initiate from a $d$-dimensional data distribution $\mathbf{x}_0 \sim q_{\text{data}}$ and define a diffusion process via a forward stochastic differential equation (SDE) of the form:
\begin{equation}\label{diffusion_sde}
\mathrm{d} \mathbf{x}_t = \mathbf{f}(\mathbf{x}_t, t) \mathrm{d} t+g_t \mathrm{d} \boldsymbol{w}_t,
\end{equation}
where $t \in [0, T]$, $\mathbf{f}: \mathbb{R}^d \times [0, T] \rightarrow \mathbb{R}^d$ denotes the drift function, $g:[0, T] \rightarrow \mathbb{R}$ is the diffusion coefficient, and $\boldsymbol{w}_t \in \mathbb{R}^d$ represents a standard Wiener process. To enable transitions between arbitrary distributions, DDBMs incorporate Doob’s $h$-transform \cite{ASDE}, a mathematical tool that modifies the drift of the forward SDE such that the process passes through a predefined terminal state $\mathbf{x}_T \in \mathbb{R}^d$. After applying Doob’s $h$-transform, the forward diffusion bridge process becomes:
\begin{equation}\label{doob}
\mathrm{d} \mathbf{x}_t = \left[ \mathbf{f}(\mathbf{x}_t, t) + g^2_t \mathbf{h}(\mathbf{x}_t, t, \mathbf{x}_T, T) \right] \mathrm{d} t+g_t \mathrm{d} \boldsymbol{w}_t,
\end{equation}
where $\mathbf{h}(\mathbf{x}_t, t, \mathbf{x}_T, T) = \nabla_{\mathbf{x}_t} \log p(\mathbf{x}_T \mid \mathbf{x}_t)$ is the $h$-function. This formulation allows the diffusion bridge to connect any initial state $\mathbf{x}_0$ to a specified terminal $\mathbf{x}_T$, making it particularly suitable for image restoration and translation applications. 
% The corresponding reverse-time SDE \cite{ANDERSON1982313} for the diffusion bridge is given by:
% \begin{equation}\label{reverse-bridge-sde}
% \begin{aligned}
% \mathrm{d} \mathbf{x}_t = & \Big[ \mathbf{f}(\mathbf{x}_t, t) + g^2_t \nabla_{\mathbf{x}_t} \log p(\mathbf{x}_T \mid \mathbf{x}_t) \\
% & - g_t^2 \nabla_{\mathbf{x}_t} \log p(\mathbf{x}_t \mid \mathbf{x}_T) \Big] \mathrm{d} t+g_t \mathrm{d} \tilde{\boldsymbol{w}}_t,
% \end{aligned}
% \end{equation}
% where $\tilde{\boldsymbol{w}}_t$ denotes the reverse-time Wiener process. The unknown term $\nabla_{\mathbf{x}_t} \log p(\mathbf{x}_t \mid \mathbf{x}_T)$ can be approximated using a score-based neural network $s_{\theta}$ \cite{score-based}.

\subsection{Generalized Ornstein-Uhlenbeck Bridge}
Generalized Ornstein-Uhlenbeck (GOU) process is a mean-reverting stochastic model widely applied in finance, physics, and related disciplines. It is characterized by the following stochastic differential equation (SDE) form \cite{introsde, Pavliotis2014, WANG2018921}:
\begin{equation}\label{gou_process}
\mathrm{d} \mathbf{x}_t=\theta_t\left(\boldsymbol{\mu}-\mathbf{x}_t\right) \mathrm{d} t+g_t \mathrm{d} \boldsymbol{w}_t,
\end{equation}
where $\boldsymbol{\mu}$ is a given state vector, $\theta_t$ denotes a scalar drift coefficient and $g_t$ represents the diffusion coefficient with $\theta_t$, $g_t$ satisfying the specified relationship $g_{t}^{2} = 2 \lambda^2 \theta_t$ where $\lambda^2$ is a given constant scalar. Built upon this structure, Generalized Ornstein–Uhlenbeck Bridge (GOUB) \cite{GOUB} constitutes a diffusion bridge framework applicable to image restoration without the need of prior knowledge. In this setup, the initial state $\mathbf{x}_0$ corresponds to a high-quality image, while the terminal state $\mathbf{x}_T = \boldsymbol{\mu}$ represents its degraded counterpart. Through Doob's $h$-transform, denote $\bar{\theta}_{s:t} = \int_{s}^{t} \theta_z dz$, $\bar{\theta}_{t} = \int_{0}^{t} \theta_z dz$ for simplification when $s=0$ and $\bar{\sigma}^2_{s:t} = \lambda^2(1-e^{-2\bar{\theta}_{s:t}})$, the forward transition $p(\mathbf{x}_t\mid\mathbf{x}_0,\mathbf{x}_T)$ of GOUB is given by
\begin{equation}\label{gou_transition}
\begin{gathered}
p(\mathbf{x}_t\mid\mathbf{x}_0,\mathbf{x}_T)=\mathcal{N}(\bar{\boldsymbol{\mu}}_{t}^{\prime},\bar{\sigma}_t^{\prime2}\mathbf{I}), \\
\bar{\boldsymbol{\mu}}_{t}^{\prime}=e^{-\bar{\theta}_t}\frac{\bar{\sigma}_{t:T}^2}{\bar{\sigma}_T^2}\mathbf{x}_0 + \left( 1 - e^{-\bar{\theta}_t}\frac{\bar{\sigma}_{t:T}^2}{\bar{\sigma}_T^2} \right)\mathbf{x}_T, \ \bar{\sigma}_t^{\prime2}=\frac{\bar{\sigma}_t^2\bar{\sigma}_{t:T}^2}{\bar{\sigma}_T^2}.
\end{gathered}
\end{equation}

\subsection{Stochastic Optimal Control}
Stochastic Optimal Control (SOC) provides a mathematical framework for deriving optimal control policies in dynamical systems subject to uncertainty. Combining stochastic modeling with optimization principles, SOC aims to identify control strategies that perform optimally under random disturbances, with established applications in areas such as finance \cite{Geering2010} and style transfer \cite{RB}. For the SDE in \eqref{diffusion_sde}, we now consider the following Linear Quadratic SOC formulation \cite{SOCtheory, gmpsb}:
\begin{equation}\label{control_problem_sde}
\begin{gathered}
\min _{\mathbf{u}_{t, \gamma} \in \mathcal{U}} \mathbb{E}\left[\int_0^T \frac{1}{2}\left\|\mathbf{u}_{t, \gamma}\right\|_2^2 d t+\frac{\gamma}{2}\left\|\mathbf{x}_T^u-x_T\right\|_2^2\right] \\
\text{s.t.} \ \mathrm{d} \mathbf{x}_t = [ \mathbf{f}(\mathbf{x}_t, t) + g_t \mathbf{u}_{t, \gamma} ] \mathrm{d} t + g_t \mathrm{d} \boldsymbol{w}_t, \ \mathbf{x}_0^u=x_0,
\end{gathered}
\end{equation}
where $\mathbf{x}_t^u$ is the diffusion process under control, $x_0$ and $x_T$ represent for the initial state and the preset terminal respectively, $\left\|\mathbf{u}_{t, \gamma}\right\|_2^2$ is the instantaneous cost, $\frac{\gamma}{2}\left\|\mathbf{x}_T^{u}-x_T\right\|_2^2$ is the terminal cost with its penalty coefficient $\gamma$. The SOC problem aims to design the controller $\mathbf{u}_{t, \gamma}$ to drive the dynamic system from $x_0$ to $x_T$ with minimum cost. 

\section{UniDB Framework}
\subsection{Constructing Diffusion Bridges via SOC Problem}
The forward SDE of diffusion bridge with Doob’s $h$-transform is constrained to start from the predetermined origin $x_0$ and end at the terminal $x_T$. Pursuing a similar goal, UniDB constructs an SOC problem where the constraints consist of an arbitrary linear SDE and a given initial state, while the optimization objective incorporates a quadratic penalty term driving the forward diffusion process to approach the predetermined terminal $x_T$. Considering $\mathbf{f}(\mathbf{x}_t, t) = f_t \mathbf{x}_t + h_t \mathbf{m}$ which is a simple reformulation of the parameters in GOU process \eqref{gou_process}, our constructed SOC problem is formed as: 
\begin{equation}\label{SOC_problem_generalized_sde}
\begin{gathered}
\min _{\mathbf{u}_{t, \gamma} \in \mathcal{U}} \mathbb{E}\left[\int_0^T \frac{1}{2}\left\|\mathbf{u}_{t, \gamma}\right\|_2^2 d t+\frac{\gamma}{2}\left\|\mathbf{x}_T^u-x_T\right\|_2^2\right] \\
\text{s.t.} \ \mathrm{d} \mathbf{x}_t = [ f_t \mathbf{x}_t + h_t \mathbf{m} + g_t \mathbf{u}_{t, \gamma} ] \mathrm{d} t + g_t \mathrm{d} \boldsymbol{w}_t, \ \mathbf{x}_0^u = x_0.
\end{gathered}
\end{equation}

According to the certainty equivalence principle \cite{RB}, the addition of noise or perturbations to a linear system with quadratic costs does not change the optimal control, we can modify the SOC problem with the deterministic ODE condition to obtain the optimal controller $\mathbf{u}_{t, \gamma}^{*}$ as follows, 
\begin{equation}\label{SOC_problem_generalized_ode}
\begin{gathered}
\min_{\mathbf{u}_{t, \gamma} \in \mathcal{U}} \int_0^T \frac{1}{2}\left\|\mathbf{u}_{t, \gamma}\right\|_2^2 d t+\frac{\gamma}{2}\left\|\mathbf{x}_T^u-x_T\right\|_2^2 \\
\text{s.t.} \ \mathrm{d} \mathbf{x}_t = [ f_t \mathbf{x}_t + h_t \mathbf{m} + g_t \mathbf{u}_{t, \gamma} ] \mathrm{d} t, \ \mathbf{x}_0^u=x_0.
\end{gathered}
\end{equation}
We can derive the closed-form solution to the problem \eqref{SOC_problem_generalized_ode}, which leads to the following Theorem \ref{theorem_4.1}: 
\begin{theorem}\label{theorem_4.1} 
\textit{Consider the SOC problem \eqref{SOC_problem_generalized_ode}, denote $d_{t, \gamma} = \gamma^{-1} + e^{2\bar{f}_{T}} \bar{g}^2_{t:T}$, $\bar{f}_{s:t} = \int_{s}^{t} f_z dz$, $\bar{h}_{s:t} = \int_{s}^{t} e^{-\bar{f}_{z}} h_z dz$ and $\bar{g}^2_{s:t} = \int_{s}^{t} e^{-2\bar{f}_{z}}g^2_z dz$, denote $\bar{f}_{t}$, $\bar{h}_{t}$ and $\bar{g}^2_{t}$ for simplification when $s=0$, then the closed-form optimal controller $\mathbf{u}_{t,\gamma}^{*}$ is} 
\begin{equation}\label{general_optimal_controller}
\mathbf{u}_{t, \gamma}^{*} = g_t e^{\bar{f}_{t:T}} \frac{x_{T} - e^{\bar{f}_{t:T}} \mathbf{x}_t - \mathbf{m} e^{\bar{f}_{T}} \bar{h}_{t:T}}{d_{t, \gamma}},
\end{equation}
\textit{and the transition of $\mathbf{x}_t$ from $x_0$ and $x_T$ is}
\begin{equation}\label{general_interpolant}
\mathbf{x}_t = e^{\bar{f}_{t}} \Bigg(\frac{d_{t, \gamma}}{d_{0, \gamma}} x_0 + \frac{e^{\bar{f}_{T}} \bar{g}^2_{t}}{d_{0, \gamma}} x_T + \Big(\bar{h}_{t} - \frac{e^{2\bar{f}_{T}} \bar{h}_{T} \bar{g}^2_{t}}{d_{0, \gamma}}\Big) \mathbf{m}\Bigg). 
\end{equation}
\end{theorem}
The proof of Theorem \ref{theorem_4.1} is provided in Appendix \ref{proof_theorem_4.1}. With Theorem \ref{theorem_4.1}, we can obtain an optimally controlled forward SDE connected from $x_0$ to the neighborhood of the terminal $x_T$ and the transition of $\mathbf{x}_t$ for the forward process. 

\subsection{Diffusion Bridges via Doob's $h$-transform is A Special Case}
We can intuitively see from the SOC problem that when we take $\gamma \to \infty$ in Theorem \ref{theorem_4.1}, the target of SDE process is precisely the predetermined endpoint \cite{gmpsb}, which is also the main purpose of Doob's $h$-transform and facilitates the following theorem:
\begin{theorem}\label{theorem_4.2} 
\textit{For the SOC problem \eqref{SOC_problem_generalized_ode}, when $\gamma \to \infty$, the optimal controller becomes $\mathbf{u}^{*}_{t, \infty} = g_t \nabla_{\mathbf{x}_t} \log p(\mathbf{x}_T \mid \mathbf{x}_t)$, and the corresponding forward SDE is the same as Doob's $h$-transform as in \eqref{doob}.}
\end{theorem}
The proof of Theorem \ref{theorem_4.2} is provided in Appendix \ref{proof_theorem_4.2}. This theorem shows that existing diffusion bridge models using Doob's $h$-transform are merely special instances of our UniDB framework, which offers a more generalized approach to diffusion bridges through the lens of SOC. Furthermore, using Doob's $h$-transform in diffusion bridge models is not necessarily optimal, as letting the terminal penalty coefficient $\gamma \to \infty$ eliminates the consideration of control costs in SOC. To support this argument, we present Proposition \ref{proposition_4.3}, which asserts that the diffusion bridge with Doob's $h$-transform is not the most effective choice. 

\begin{proposition}\label{proposition_4.3} 
\textit{Consider the SOC problem \eqref{SOC_problem_generalized_ode}, denote $\mathcal{J}(\mathbf{u}_{t, \gamma}, \gamma) \triangleq \int_0^T \frac{1}{2} \left\|\mathbf{u}_{t, \gamma}\right\|_2^2 d t+\frac{\gamma}{2}\left\|\mathbf{x}_T^{u}-x_T\right\|_2^2$ as the overall cost of the system, $\mathbf{u}_{t, \gamma}^{*}$ as the optimal controller \eqref{general_optimal_controller}, then}
\begin{equation}
\mathcal{J}(\mathbf{u}_{t, \gamma}^{*}, \gamma) \le \mathcal{J}(\mathbf{u}_{t, \infty}^{*}, \infty).
\end{equation}
\end{proposition}
Detailed proof of Proposition \ref{proposition_4.3} is provided in Appendix \ref{proof_proposition_4.3}. Proposition \ref{proposition_4.3} shows that finite $\gamma$ achieves a lower total cost not by sacrificing performance, but by optimally trading minor terminal mismatches for significantly smoother and more natural diffusion paths. Doob’s $h$-transform requires larger controller $\|\mathbf{u}_{t, \infty}^{*}\|_2^2 \geq \|\mathbf{u}_{t, \gamma}^{*}\|_2^2$ in SDE trajectory to force precise endpoint matching, which may disrupt the inherent continuity and smoothness of images. As shown in Figure \ref{fig_main_experiment}, Doob's $h$-transform can lead to artifacts along edges and unnatural patterns in smooth regions. Therefore, maintaining the penalty coefficient $\gamma$ as a hyperparameter is a more effective approach.

% Our UniDB is a unified framework for existing diffusion bridge models: DDBMs (VE) \cite{DDBMs}, DDBMs (VP) \cite{DDBMs}, and GOUB \cite{GOUB}. 
% \begin{proposition}\label{proposition_4.4} UniDB encompasses existing diffusion bridge models by employing different hyper-parameter spaces $\mathcal{H}$ as follows:

% \begin{itemize}
% \item DDBMs (VE) corresponds to UniDB with hyper-parameter $\mathcal{H}_\text{VE}(f_t=0, h_t=0, \gamma \rightarrow \infty)$

% \item DDBMs (VP) corresponds to UniDB with hyper-parameter $\mathcal{H}_\text{VP}(f_t=-\frac{1}{2} g^2_t, h_t=0, \gamma \rightarrow \infty)$

% \item GOUB corresponds to UniDB with hyper-parameter $\mathcal{H}_\text{GOU}(f_t=\theta_t, h_t=-\theta_t, \mathbf{m} = \boldsymbol{\mu}, \gamma \rightarrow \infty)$
% \end{itemize}
% \end{proposition}

% Details of the proposition \ref{proposition_4.4} are provided in Appendix \ref{proof_proposition_4.4}. Evidently, these diffusion bridge models like DDBMs (VE), DDBMs (VP), and GOUB all based on Doob's $h$-transform are all special cases of UniDB with $\gamma \rightarrow \infty$. 

\subsection{An Example: UniDB-GOU}\label{example}
Our UniDB is a unified framework for existing diffusion bridge models like DDBMs (VE) \cite{DDBMs}, DDBMs (VP) \cite{DDBMs}, and GOUB \cite{GOUB} as all of them based on Doob's $h$-transform are special cases of UniDB with $\gamma \rightarrow \infty$ and employing specific hyper-parameters $f_t$, $h_t$, and $\mathbf{m}$. However, according to Proposition \ref{proposition_4.3}, these models are not the effective choices. Therefore, we introduce UniDB based on the GOU process \eqref{gou_process}, hereafter referred to as UniDB-GOU, which retains the penalty coefficient $\gamma$ as the hyper-parameter. Considering the SOC problem with GOU process \eqref{gou_process}, the optimally controlled forward SDE is:
\begin{equation}\label{forward_unidb_gou_sde}
\begin{gathered}
\mathrm{d} \mathbf{x}_t = \left[ \theta_t (x_T - \mathbf{x}_t) + g_t \mathbf{u}_{t, \gamma}^{*}\right] \mathrm{d} t + g_t \mathrm{d} \boldsymbol{w}_t, \\
\mathbf{u}_{t, \gamma}^{*} = \frac{g_t e^{-2\bar{\theta}_{t:T}}}{\gamma^{-1} + \bar{\sigma}^2_{t:T}}.
\end{gathered}
\end{equation}
and the mean value of forward transition $p(\mathbf{x}_t \mid x_0, x_T)$ is
\begin{equation}\label{forward_unidb_gou_mean_transition}
\bar{\boldsymbol{\mu}}_{t, \gamma} = \xi_t x_0 + (1 - \xi_t) x_T, \ \xi_t = e^{-\bar{\theta}_{t}} \frac{\gamma^{-1} + \bar{\sigma}^2_{t:T}}{\gamma^{-1} + \bar{\sigma}^2_{T}}.
\end{equation}
The derivation can be summarized by setting $f_t=\theta_t$, $h_t=-\theta_t$, and $\mathbf{m} = x_T$ in Theorem \ref{theorem_4.1} and simplifying. The related backward reverse SDE \cite{ANDERSON1982313} of UniDB-GOU is then formed as: 
\begin{equation}\label{unidb_gou_reverse_sde}
\mathrm{d} \mathbf{x}_t = \left[ \theta_t (x_T - \mathbf{x}_t) + g_t \mathbf{u}_{t, \gamma}^{*} + \frac{g^2_t}{\bar{\sigma}_t^{\prime}}  \boldsymbol{\epsilon}_{\theta}(\mathbf{x}_t, x_T, t)\right]  \mathrm{d} t + g_t \mathrm{d} \tilde{\boldsymbol{w}}_t,
\end{equation}
where we take the parameterization $\nabla_{\mathbf{x}_t} \log p(\mathbf{x}_t \mid x_T) \approx -\boldsymbol{\epsilon}_{\theta}(\mathbf{x}_t, x_T, t)/\bar{\sigma}_t^{\prime}$. For training objectives, UniDB can be adapted based on different existing diffusion bridges. Whether maximum log-likelihood \cite{DDPM, GOUB} and conditional score matching \cite{score-based, DDBMs}, the training objective is based on the forward transition $p(\mathbf{x}_t\mid\mathbf{x}_0,\mathbf{x}_T)$. Thus, we need to derive this probability. The closed-form expression in \eqref{forward_unidb_gou_mean_transition} represents the mean value of the forward transition after applying reparameterization techniques. However, this expression lacks a noise component after the transformation based on the certainty equivalence principle. To address this issue, we employ stochastic interpolant theory \cite{SI} to introduce the same noise term $\bar{\sigma}_t^{\prime2}$ as in \eqref{gou_transition}, leading to the following forward transition: \begin{equation}\label{forward_unidb_gou_transition}
\begin{gathered}
p(\mathbf{x}_t\mid x_0, x_T)=\mathcal{N}(\bar{\boldsymbol{\mu}}_{t, \gamma},\bar{\sigma}_t^{\prime2}\mathbf{I}).
\end{gathered}
\end{equation}

\begin{algorithm}[t]
   \caption{UniDB Sampling}
   \label{sampling_pseudo_code}
\begin{algorithmic}
    \STATE {\bfseries Input:} Low-Quality images $x_T$ and noise prediction model $\boldsymbol{\epsilon}_{\theta}(\mathbf{x}_t, x_T, t)$. Initialize $\mathbf{x}_T = x_T$.
    \FOR{$t=T$ {\bfseries to} $1$}
        \STATE Sample $\boldsymbol{z} \sim \mathcal{N}(0, I)$ if $t > 1$, else $z = 0$.
        \STATE $\hat{\boldsymbol{\epsilon}}_t = \boldsymbol{\epsilon}_{\theta}(\mathbf{x}_t, x_T, t)$.
        \STATE $\mathbf{x}_{t-1} = \mathbf{x}_{t} - ( \theta_t + g^2_t \frac{e^{-2\bar{\theta}_{t:T}}}{\gamma^{-1} + \bar{\sigma}^2_{t:T}}) (x_T - \mathbf{x}_t) + \frac{g^2_t}{\bar{\sigma}_{t}^{\prime}} \hat{\boldsymbol{\epsilon}}_t - g_t \boldsymbol{z}$.
   \ENDFOR
   \STATE \textbf{Return} High-Quality images $\tilde{\mathbf{x}}_0$.
\end{algorithmic}
\end{algorithm}

\begin{algorithm}[t]
   \caption{UniDB++ Sampling ($k=1$)}
   \label{unidb_data_sde_solver_1_sampling}
\begin{algorithmic}
    \STATE {\bfseries Input:} Low-Quality images $x_T$, data prediction model $\mathbf{x}_\theta (\mathbf{x}_{t}, x_T, t)$, and $M+1$ time steps $\left\{t_i\right\}_{i=0}^M$ decreasing from $t_0=T$ to $t_M=0$. Initialize $\mathbf{x}_{t_0} = x_T$.
    \FOR{$i=1$ {\bfseries to} $M$}
        \STATE Sample $\boldsymbol{z} \sim \mathcal{N} (0, I)$ if $i < M$, else $\boldsymbol{z} = 0$.
        \STATE $\hat{\mathbf{x}}_{0|t_{i-1}} = \mathbf{x}_\theta(\mathbf{x}_{t_{i-1}}, \mathbf{x}_T, t_{i-1})$.
        \STATE $\mathbf{x}_{t_{i}} =  \frac{\kappa_{t_{i}, \gamma} \kappa_{t_{i-1}} \rho_{t_{i}}}{\kappa_{t_{i-1}, \gamma} \kappa_{t_{i}} \rho_{t_{i-1}}} \mathbf{x}_{t_{i-1}} + \Big( 1 - \frac{\kappa_{t_{i}, \gamma} \kappa_{t_{i-1}} \rho_{t_{i}}}{\kappa_{t_{i-1}, \gamma} \kappa_{t_{i}} \rho_{t_{i-1}}} + \frac{\kappa_{t_{i}, \gamma} \kappa_{t_{i-1}} \rho_{t_{i}}}{\kappa_{0, \gamma} \kappa_{t_{i}} \rho_{t_{i-1}}} - \frac{\kappa_{t_{i}, \gamma}}{\kappa_{0, \gamma}} \Big) x_T + \Big(\frac{\kappa_{t_{i}, \gamma}}{\kappa_{0, \gamma}} - \frac{\kappa_{t_{i}, \gamma} \kappa_{t_{i-1}} \rho_{t_{i}}}{\kappa_{0, \gamma} \kappa_{t_{i}} \rho_{t_{i-1}}}\Big) \hat{\mathbf{x}}_{0|t_{i-1}} + \delta^{d}_{t_{i-1}:t_{i}, \gamma} \boldsymbol{z}$.
   \ENDFOR
   \STATE \textbf{Return} High-Quality images $\tilde{\mathbf{x}}_0$.
\end{algorithmic}
\end{algorithm}

In terms of the sampling procedure, we can recover or generate the origin image $\tilde{\mathbf{x}}_0$ through naive Euler sampling iterations of the reverse-time SDE \eqref{unidb_gou_reverse_sde} as in Algorithm \ref{sampling_pseudo_code}.

\section{UniDB++ for Fast Sampling}
\subsection{Solutions to the Reverse-Time SDE with Data Prediction}

Since the naive Euler sampling method typically requires hundreds of sampling steps to achieve merely acceptable image restoration or generation quality which significantly limits its practical efficiency, to address this limitation, we focus on deriving the exact closed-form solutions for UniDB-GOU’s reverse-time SDE. 

Notably, some prior SDE solvers for diffusion models based on the noise prediction model $\boldsymbol{\epsilon}_{\theta}(\mathbf{x}_t, x_T, t)$ often demonstrated significant numerical instability, especially under low sampling step conditions, which has been justified through many ablation experiments in \cite{dpm-solver-plus, MaRS}. An empirical explanation is that data prediction $\mathbf{x}_{\theta}(\mathbf{x}_t, x_T, t)$ directly estimates the fixed and smooth target $\mathbf{x}_0$, ensuring stable predictions of the solvers even with few sampling steps. In contrast, noise prediction $\boldsymbol{\epsilon}_{\theta}(\mathbf{x}_t, x_T, t)$ targets dynamically changing noise $\boldsymbol{\epsilon}$, where small sampling steps can destabilize high-frequency components, leading to error accumulation. The errors of solvers with data prediction are less severe, as $\mathbf{x}_0$’s structural information remains robust, preserving global coherence. 

\begin{algorithm}[t]
   \caption{UniDB++ Sampling with Corrector ($k=1$)}
   \label{unidb_data_sde_solver_1_corrector_sampling}
\begin{algorithmic}
    \STATE {\bfseries Input:} Low-Quality images $x_T$, data prediction model $\mathbf{x}_\theta (\mathbf{x}_{t}, x_T, t)$, a buffer $Q$, and $M+1$ time steps $\left\{t_i\right\}_{i=0}^M$ decreasing from $t_0=T$ to $t_M=0$. Initialize $\mathbf{x}^{c}_{t_0} = x_T$. 
    \FOR{$i=1$ {\bfseries to} $M$}
        \STATE Sample $\boldsymbol{z}_1, \boldsymbol{z}_2 \sim \mathcal{N} (0, I)$ if $i < M$, else $\boldsymbol{z}_1 = \boldsymbol{z}_2 = 0$.
        \STATE Take $\hat{\mathbf{x}}_{0|i-1} = \mathbf{x}_\theta(\mathbf{x}_{t_{i-1}}, x_T, t_{i-1})$ from $Q$.
        \STATE $\mathbf{x}_{t_{i}} = \frac{\kappa_{t_{i}, \gamma} \kappa_{t_{i-1}} \rho_{t_{i}}}{\kappa_{t_{i-1}, \gamma} \kappa_{t_{i}} \rho_{t_{i-1}}} \mathbf{x}^{c}_{t_{i-1}} + \Big( 1 - \frac{\kappa_{t_{i}, \gamma} \kappa_{t_{i-1}} \rho_{t_{i}}}{\kappa_{t_{i-1}, \gamma} \kappa_{t_{i}} \rho_{t_{i-1}}} + \frac{\kappa_{t_{i}, \gamma} \kappa_{t_{i-1}} \rho_{t_{i}}}{\kappa_{0, \gamma} \kappa_{t_{i}} \rho_{t_{i-1}}} - \frac{\kappa_{t_{i}, \gamma}}{\kappa_{0, \gamma}} \Big) x_T + (\frac{\kappa_{t_{i}, \gamma}}{\kappa_{0, \gamma}} - \frac{\kappa_{t_{i}, \gamma} \kappa_{t_{i-1}} \rho_{t_{i}}}{\kappa_{0, \gamma} \kappa_{t_{i}} \rho_{t_{i-1}}}) \hat{\mathbf{x}}_{0|i-1} + \delta^{d}_{t_{i-1}:t_{i}, \gamma} \boldsymbol{z}_1$. % \COMMENT{UniDB++ ($k=1$)}
        \STATE $\hat{\mathbf{x}}_{0|t_{i}} = \mathbf{x}_\theta(\mathbf{x}_{t_{i}}, x_T, t_{i})$ and $Q \leftarrow \hat{\mathbf{x}}_{0|t_{i}}$.
        \STATE $\mathbf{x}^{c}_{t_{i}} = \frac{\kappa_{t_{i}, \gamma} \kappa_{t_{i-1}} \rho_{t_{i}}}{\kappa_{t_{i-1}, \gamma} \kappa_{t_{i}} \rho_{t_{i-1}}} \mathbf{x}^{c}_{t_{i-1}} + \Big( 1 - \frac{\kappa_{t_{i}, \gamma} \kappa_{t_{i-1}} \rho_{t_{i}}}{\kappa_{t_{i-1}, \gamma} \kappa_{t_{i}} \rho_{t_{i-1}}} + \frac{\kappa_{t_{i}, \gamma} \kappa_{t_{i-1}} \rho_{t_{i}}}{\kappa_{0, \gamma} \kappa_{t_{i}} \rho_{t_{i-1}}} - \frac{\kappa_{t_{i}, \gamma}}{\kappa_{0, \gamma}} \Big) x_T + (\frac{\kappa_{t_{i}, \gamma}}{\kappa_{0, \gamma}} - \frac{\kappa_{t_{i}, \gamma} \kappa_{t_{i-1}} \rho_{t_{i}}}{\kappa_{0, \gamma} \kappa_{t_{i}} \rho_{t_{i-1}}}) \hat{\mathbf{x}}_{0|i-1} + B(h_i)\frac{\kappa_{t, \gamma}c_i}{\kappa_{0, \gamma}} (\hat{\mathbf{x}}_{0|t_{i}} - \hat{\mathbf{x}}_{0|t_{i-1}}) + \delta^{d}_{t_{i-1}:t_{i}, \gamma} \boldsymbol{z}_2$.
   \ENDFOR
   \STATE \textbf{Return} High-Quality images $\tilde{\mathbf{x}}^{c}_0$.
\end{algorithmic}
\end{algorithm}

Inspired by this, we start by reformulating the reverse-time SDE from the noise prediction model to the data prediction model and analyzing its exact solution to accelerate the sampling procedure \eqref{unidb_gou_reverse_sde} \cite{dpm-solver, dpm-solver-plus, MaRS}. Following the transition probability proposed in \eqref{forward_unidb_gou_transition}, we can introduce the data prediction model $\mathbf{x}_{\theta}(\mathbf{x}_t, x_T, t)$ which predicts the original data $\mathbf{x}_0$ based on the noisy $\mathbf{x}_t$ \cite{EDM, dpm-solver-plus} as
\begin{equation}\label{relation_data_noise}
\mathbf{x}_{\theta}(\mathbf{x}_t, x_T, t) = \frac{1}{\xi_t}(\mathbf{x}_t - (1 - \xi_t)x_T - \bar{\sigma}_{t}^{\prime}\boldsymbol{\epsilon}_{\theta}\left(\mathbf{x}_t, x_T, t)\right).
\end{equation}
Combining \eqref{relation_data_noise} with \eqref{unidb_gou_reverse_sde}, we can derive the reverse SDE w.r.t. data prediction model, respectively: 
\begin{equation}\label{unidb_gou_reverse_sde_data}
\begin{aligned}
\mathrm{d} \mathbf{x}_t = \Bigg[&\left(\theta_t - \frac{g_t^2(1-\xi_t)}{\bar{\sigma}_{t}^{\prime2}}\right)x_T-\left(\theta_t - \frac{g_t^2}{\bar{\sigma}_{t}^{\prime2}} \right)\mathbf{x}_t \\
&+ g_t\mathbf{u}_{t, \gamma}^{*}- \frac{g_t^2\xi_t}{\bar{\sigma}_{t}^{\prime2}} \mathbf{x}_\theta(\mathbf{x}_t, x_T, t) \Bigg] \mathrm{d} t + g_t \mathrm{d} \tilde{\boldsymbol{w}}_t,
\end{aligned}
\end{equation}
% \begin{equation}\label{unidb_gou_reverse_mean_ode_data}
% \begin{aligned}
% \mathrm{d} \mathbf{x}_t = \Bigg[&\left(\theta_t + \frac{g_t^2(1-\xi_t)}{\bar{\sigma}_{t}^{\prime2}}\right)x_T-(\theta_t - \frac{g_t^2}{\bar{\sigma}_{t}^{\prime2}} )\mathbf{x}_t \\
% &+ g_t\mathbf{u}_{t, \gamma}^{*}- \frac{g_t^2\xi_t}{\bar{\sigma}_{t}^{\prime2}} \mathbf{x}_\theta(\mathbf{x}_t, x_T, t) \Bigg] \mathrm{d} t.
% \end{aligned}
% \end{equation}

\begin{table*}[t]
  \centering
  \caption{Quantitative comparison between both UniDB, UniDB++ and the relevant baselines on different image restoration tasks.}
  \label{table_main_sde_all}
  \vspace{-3mm}
  \fontsize{10pt}{12pt}\selectfont
  \resizebox{0.84\textwidth}{!}{
  \begin{tabular}{cccccccccccccc}
    \toprule
    & & \multicolumn{4}{c}{\textbf{Image 4$\times$Super-Resolution}} & \multicolumn{4}{c}{\textbf{Image Deraining}}\\
            \cmidrule(lr){3-6} \cmidrule(lr){7-10} 
    \textbf{Method} & \textbf{NFE$\downarrow$} & \textbf{PSNR$\uparrow$} & \textbf{SSIM$\uparrow$} & \textbf{LPIPS$\downarrow$} & \textbf{FID$\downarrow$} & \textbf{PSNR$\uparrow$} & \textbf{SSIM$\uparrow$} & \textbf{LPIPS$\downarrow$} & \textbf{FID$\downarrow$} \\
    \midrule
    IR-SDE & 100 & 25.90 & 0.6570 & 0.231 & 45.36 & 31.65 & 0.9040 & 0.047 & 18.64 \\
    DDBMs & 100 & 24.21 & 0.5808 & 0.384 & 36.55 & 30.89 & 0.8847 & 0.051 & 23.36 \\
    GOUB & 100 & 26.89 & 0.7478 & 0.220 & 20.85 & 31.96 & 0.9028 & 0.046 & 18.14 \\ 
    % \midrule
    MaRS & 5 & 27.73 & 0.7831 & 0.286 & 29.20 & 27.60 & 0.9087 & 0.070 & 28.44 \\
    DBIMs & 5 & 28.05 & 0.7950 & 0.260 & 20.12 & 33.13 & 0.9219 & 0.056 & 21.09 \\
    \midrule
    UniDB (ours) & 100 & 25.46 & 0.6856 & 0.179 & 16.21 & 32.05 & 0.9036 & 0.045 & 17.65 \\
    UniDB++ (ours) & 5 & \textbf{28.40} & \textbf{0.8045} & 0.235 & 17.06 & \textbf{33.26} & \textbf{0.9340} & 0.050 & 18.92 \\
    UniDB++ (ours) & 20 & 27.38 & 0.7773 & 0.179 & 15.26 & 32.57 & 0.9225 & 0.045 & 17.54 \\
    UniDB++ (ours) & 50 & 26.61 & 0.7535 & \textbf{0.159} & \textbf{14.98} & 32.17 & 0.9158 & \textbf{0.043} & \textbf{17.19} \\
    \midrule
    & & \multicolumn{4}{c}{\textbf{Image Inpainting}} & \multicolumn{4}{c}{\textbf{Image Raindrop Restoration}} \\
            \cmidrule(lr){3-6} \cmidrule(lr){7-10} 
    \textbf{Method} & \textbf{NFE$\downarrow$} & \textbf{PSNR$\uparrow$} & \textbf{SSIM$\uparrow$} & \textbf{LPIPS$\downarrow$} & \textbf{FID$\downarrow$} & \textbf{PSNR$\uparrow$} & \textbf{SSIM$\uparrow$} & \textbf{LPIPS$\downarrow$} & \textbf{FID$\downarrow$} \\
    \midrule
    IR-SDE & 100 & 27.43 & 0.8937 & 0.051 & 7.54 & 26.87 & 0.7896 & 0.086 & 33.82 \\
    DDBMs & 100 & 27.32 & 0.8841 & 0.049 & 11.87 & 29.22 & 0.8731 & 0.078 & 29.49 \\ % & 25.67 & 0.5519 & 0.239 & 55.33 \\
    GOUB & 100 & 28.98 & 0.9067 & 0.037 & 4.30 & 27.69 & 0.8266 & 0.077 & 36.70 \\ 
    % \midrule
    MaRS & 10 & 28.07 & 0.9129 & 0.047 & 6.94 & 29.29 & 0.8872 & 0.079 & 32.62 \\
    DBIMs & 10 & 28.19 & 0.9178 & 0.053 & 8.58 & 29.48 & \textbf{0.9053} & 0.064 & 32.46 \\
    \midrule
    UniDB (ours) & 100 & 29.20 & 0.9077 & \textbf{0.036} & 4.08 & 28.61 & 0.8162 & 0.069 & 26.07 \\
    UniDB++ (ours) & 10 & 28.45 & \textbf{0.9230} & 0.045 & 6.08 & \textbf{30.66} & 0.9027 & 0.067 & 27.05 \\
    UniDB++ (ours) & 20 & \textbf{29.31} & 0.9206 & 0.038 & 4.28 & 30.22 & 0.8915 & 0.065 & 24.95 \\
    UniDB++ (ours) & 50 & 29.18 & 0.9161 & \textbf{0.036} & \textbf{3.88} & 29.64 & 0.8754 & \textbf{0.062} & \textbf{24.37} \\
    \bottomrule
  \end{tabular}
  }
\end{table*}

To streamline computational complexity and avoid symbol conflict, we adopt a modified parameterization scheme that differs from prior implementations \cite{dpm-solver, MaRS}. Specifically, define $\kappa_{t} = e^{\bar{\theta}_{t: T}}(1 - e^{-2 \bar{\theta}_{t: T}})$, $\kappa_{t, \gamma} = e^{\bar{\theta}_{t: T}}((\gamma \lambda^2)^{-1} + 1 - e^{-2 \bar{\theta}_{t: T}})$, $\rho_t = e^{\bar{\theta}_{t}}(1 - e^{-2 \bar{\theta}_{t}})$ and $\beta_{t} = \log (\kappa_{t}/\rho_t)$ similar to the half log-SNR (signal-to-noise-ratio) $\lambda_t$ in \cite{dpm-solver}. With a simple verification revealing that $\beta_{t, \gamma}$ is a strictly decreasing function of time $t$ so it has an inverse function $t_\beta(\cdot)$ satisfying $t = t_\beta(\beta_t)$, we can further change the subscripts of $\mathbf{x}_t$ to $\mathbf{x}_\beta$ in the data prediction model as $\hat{\mathbf{x}}_\theta\left(\mathbf{x}_\beta, x_T,\beta\right):=\mathbf{x}_\theta\left(\mathbf{x}_{t_\beta(\beta)}, x_T, t_\beta(\beta)\right)$ which helps simplify the formulation \cite{dpm-solver}, then solving the exact solution of the SDE \eqref{unidb_gou_reverse_sde_data}, which leads to the following theorem: 
\begin{theorem}\label{theorem_exact_solution}
Given an initial value $\mathbf{x}_s$ at time $s \in [0, T]$, the exact solution $\mathbf{x}_t$ at time $t \in [0, s]$ of the reverse-time SDE \eqref{unidb_gou_reverse_sde_data} is: 
\begin{equation}\label{sde_solution}
\begin{gathered}
\begin{aligned}
\mathbf{x}_t = \frac{\kappa_{t, \gamma} \kappa_{s} \rho_t}{\kappa_{s, \gamma} \kappa_{t} \rho_s}&\mathbf{x}_s + \left( 1 - \frac{\kappa_{t, \gamma} \kappa_{s} \rho_t}{\kappa_{s, \gamma} \kappa_{t} \rho_s} + \frac{\kappa_{t, \gamma} \rho_t \kappa_{s}}{\kappa_{0, \gamma} \kappa_{t} \rho_s} - \frac{\kappa_{t, \gamma}}{\kappa_{0, \gamma}}\right)x_T \\
&+ \frac{\kappa_{t, \gamma} \rho_t}{\kappa_{0, \gamma} \kappa_{t}} \int_{\beta_{s}}^{\beta_{t}} e^{\beta} \hat{\mathbf{x}}_\theta(\mathbf{x}_\beta, x_T, \beta) d\beta + \delta^{d}_{s:t, \gamma} \boldsymbol{z},    
\end{aligned} \\
\begin{aligned}
(\delta^{d}_{s:t, \gamma})^2 = &\frac{\lambda^2 \kappa^2_{t, \gamma}\rho^2_t}{\kappa^2_{t}} \Bigg[E \frac{e^{2\bar{\theta}_{s}} - e^{2\bar{\theta}_{t}}}{(e^{2\bar{\theta}_{t}}-1)(e^{2\bar{\theta}_{s}}-1)} \\
&- D\log \frac{\kappa_{s, \gamma}\rho_t}{\kappa_{t, \gamma}\rho_s} - F( \frac{e^{-\bar{\theta}_{T}-\bar{\theta}_{t}}}{\kappa_{t, \gamma}} - \frac{e^{-\bar{\theta}_{T}-\bar{\theta}_{s}}}{\kappa_{s, \gamma}})\Bigg],
\end{aligned}
\end{gathered}
\end{equation}
where $c_1 = (\gamma\lambda^2)^{-1}e^{2\bar{\theta}_{T}}$, $c_2 = e^{2\bar{\theta}_{T}} - 1$, $D = -2c_1c_2/(c_1 + c_2)^3$, $E=c_2^2/(c_1 + c_2)^2$, $F=c_1^2/(c_1 + c_2)^2$, and $\boldsymbol{z} \sim \mathcal{N} (0, I)$ is a standard Guassian noise.
\end{theorem}

Please refer to Appendix \ref{appendix_data_solver} for detailed derivation. Theorem \ref{theorem_exact_solution} yields a superior solution of the SDE \eqref{unidb_gou_reverse_sde_data} as it enables exact computation of the linear part about $\mathbf{x}_t$ and $x_T$, leaving only integral with the data prediction model (neural network) term subject to approximation. To calculate this integral, we follow and apply the exponential integrator method \cite{dpm-solver} and the $k$-th order Taylor expansion of $\hat{\mathbf{x}}_\theta(\mathbf{x}_\beta, x_T, \beta)$, specifically: 
\begin{equation}
\begin{aligned}
& \int_{\beta_{s}}^{\beta_{t}} e^{\beta} \hat{\mathbf{x}}_\theta(\mathbf{x}_\beta, x_T, \beta) d\beta = \mathcal{O}(h^{k+1})\\
&+ \sum_{i=0}^{k-1} e^{\beta_{t}} (\sum_{j=0}^{i} \frac{(-1)^{i-j} h^j}{j!} - (-1)^{i} e^{-h} )\hat{\mathbf{x}}_{\theta}^{(i)}(\mathbf{x}_{\beta_s}, x_T, \beta_s),
\end{aligned}
\end{equation}
where $\hat{\mathbf{x}}_{\theta}^{(i)}(\mathbf{x}_{\beta}, x_T, \beta) = \mathrm{d}^i \hat{\mathbf{x}}_{\theta}(\mathbf{x}_{\beta}, x_T, \beta)/\mathrm{d} \beta^i$ is the $i$-th order derivatives of $\hat{\mathbf{x}}_{\theta}(\mathbf{x}_{\beta}, x_T, \beta)$ w.r.t. $\beta$, $h = \beta_{t} - \beta_{s}$, and $\mathcal{O}(h^{k+1})$ is the high-order error term that can be omitted \cite{dpm-solver-plus, MaRS}. Here we demonstrate the first-order case ($k=1$) and its pseudo-code Algorithm \ref{unidb_data_sde_solver_1_sampling} of the sampling procedure for the reverse-time SDE \eqref{sde_solution} as an example:
\begin{equation}\label{unidb_1st_order_solution}
\begin{aligned}
\mathbf{x}_t = &\frac{\kappa_{t, \gamma} \kappa_{s} \rho_t}{\kappa_{s, \gamma} \kappa_{t} \rho_s}\mathbf{x}_s + \left(1 - \frac{\kappa_{t, \gamma} \kappa_{s} \rho_t}{\kappa_{s, \gamma} \kappa_{t} \rho_s} + \frac{\kappa_{t, \gamma} \rho_t \kappa_{s}}{\kappa_{0, \gamma} \kappa_{t} \rho_s} - \frac{\kappa_{t, \gamma}}{\kappa_{0, \gamma}}\right)x_T \\
&+ \left(\frac{\kappa_{t, \gamma}}{\kappa_{0, \gamma}} - \frac{\kappa_{t, \gamma} \rho_t \kappa_{s}}{\kappa_{0, \gamma} \kappa_{t} \rho_s} \right) \mathbf{x}_\theta(\mathbf{x}_{s}, x_T, s) + \delta^{d}_{s:t, \gamma} \boldsymbol{z}.
\end{aligned}
\end{equation}

In practice, to approximate the high-order derivatives, taking second-order derivative as an example, we provide two kinds of algorithm implementations: both the forward and backward difference methods, named as the single-step and multi-step methods \cite{dpm-solver-plus}, respectively. For higher-order algorithms, please refer to Algorithm \ref{unidb_data_sde_solver_2_single_step} and Algorithm \ref{unidb_data_sde_solver_2_multi_step} in Appendix \ref{appendix_updating_rule}.

% \textbf{Remark 2.} Since we have mentioned the reverse-time Mean-ODE process \eqref{unidb_gou_reverse_mean_ode_data} is constructed by directly neglecting the Brownian term of the reverse-time SDE \eqref{unidb_gou_reverse_sde}, so we can easily derive UniDB++ for the Mean-ODE process with the data prediction model by setting $\delta^{d}_{t:s} = 0$ in \eqref{sde_solution} without any cost during the sampling procedure. Specifically, the exact solution of the Mean-ODE process \eqref{unidb_gou_reverse_mean_ode_data} is 
% \begin{equation}\label{mean_ode_solution}
% \begin{aligned}
% \mathbf{x}_t = \frac{\kappa_{t, \gamma} \kappa_{s} \rho_t}{\kappa_{s, \gamma} \kappa_{t} \rho_s}\mathbf{x}_s + &\left( 1 - \frac{\kappa_{t, \gamma} \kappa_{s} \rho_t}{\kappa_{s, \gamma} \kappa_{t} \rho_s} + \frac{\kappa_{t, \gamma} \rho_t \kappa_{s}}{\kappa_{0, \gamma} \kappa_{t} \rho_s} - \frac{\kappa_{t, \gamma}}{\kappa_{0, \gamma}}\right)x_T \\
% &+ \frac{\kappa_{t, \gamma} \rho_t}{\kappa_{0, \gamma} \kappa_{t}} \int_{\beta_{s}}^{\beta_{t}} e^{\beta} \hat{\mathbf{x}}_\theta(\mathbf{x}_\beta, x_T, \beta) d\beta.    
% \end{aligned}
% \end{equation}

\begin{table*}[t]
  \centering
  \caption{Ablation Study on penalty coefficients. Quantitative comparison with different penalty coefficients $\gamma$.}
  \label{table_ablation_gamma}
  \vspace{-3mm}
  \fontsize{12pt}{14pt}\selectfont
  \resizebox{0.98\textwidth}{!}{
  \begin{tabular}{cccccccccccccc}
    \toprule
    & & \multicolumn{4}{c}{\textbf{Image 4$\times$Super-Resolution}} & \multicolumn{4}{c}{\textbf{Image Deraining}} & \multicolumn{4}{c}{\textbf{Image Inpainting}} \\
            \cmidrule(lr){3-6} \cmidrule(lr){7-10} \cmidrule(lr){11-14} 
    \textbf{Method} & \textbf{Penalty $\gamma$} & \textbf{PSNR$\uparrow$} & \textbf{SSIM$\uparrow$} & \textbf{LPIPS$\downarrow$} & \textbf{FID$\downarrow$} & \textbf{PSNR$\uparrow$} & \textbf{SSIM$\uparrow$} & \textbf{LPIPS$\downarrow$} & \textbf{FID$\downarrow$} & \textbf{PSNR$\uparrow$} & \textbf{SSIM$\uparrow$} & \textbf{LPIPS$\downarrow$} & \textbf{FID$\downarrow$} \\
    \midrule
    GOUB & $\infty$ & \textbf{26.89} & \textbf{0.7478} & 0.220 & 20.85 & 31.96 & 0.9028 & 0.046 & 18.14 & 28.98 & 0.9067 & 0.037 & 4.30 \\
    \midrule
    \multirow{3}{*}{Ours} & $1\times10^6$ & 24.72 & 0.6587 & 0.199 & 18.37 & 31.96 & 0.9018 & \textbf{0.045} & 18.37 & 29.15 & 0.9068 & \textbf{0.036} & 4.12 \\
    & $1\times10^7$ & 25.46 & 0.6856 & \textbf{0.179} & \textbf{16.21} & 32.00 & 0.9029 & 0.046 & 17.87 & \textbf{29.20} & \textbf{0.9077} & \textbf{0.036} & \textbf{4.08} \\
    & $1\times10^8$ & 25.06 & 0.6393 & 0.289 & 23.76 & \textbf{32.05} & \textbf{0.9036} & \textbf{0.045} & \textbf{17.65} & 28.65 & 0.9062 & 0.039 & 4.64 \\
    % \midrule
    % & & \multicolumn{4}{c}{\textbf{Image Inpainting}} & \multicolumn{4}{c}{\textbf{RainDrop Image Restoration}} \\
    %         \cmidrule(lr){3-6} \cmidrule(lr){7-10} 
    % \textbf{Method} & \textbf{Penalty Coefficients $\gamma$} & \textbf{PSNR$\uparrow$} & \textbf{SSIM$\uparrow$} & \textbf{LPIPS$\downarrow$} & \textbf{FID$\downarrow$} & \textbf{PSNR$\uparrow$} & \textbf{SSIM$\uparrow$} & \textbf{LPIPS$\downarrow$} & \textbf{FID$\downarrow$} \\
    % \midrule
    % GOUB & $\infty$ & 28.98 & 0.9067 & 0.037 & 4.30 & 27.69 & 0.8266 & 0.077 & 36.70 \\
    % \midrule
    % \multirow{3}{*}{Ours} & $1\times10^6$ & 29.15 & 0.9068 & \textbf{0.036} & 4.12 & \textbf{28.61} & 0.8162 & 0.070 & \textbf{26.07} \\
    % & $1\times10^7$ & \textbf{29.20} & \textbf{0.9077} & \textbf{0.036} & \textbf{4.08} & 28.28 & \textbf{0.8496} & \textbf{0.064} & 32.08 \\
    % & $1\times10^8$ & 28.65 & 0.9062 & 0.039 & 4.64 & 27.78 & 0.7846 & 0.085 & 30.56 \\
    \bottomrule
  \end{tabular}
  }
\vspace{-3mm}
\end{table*}

\begin{table*}[htbp]
    \centering
    \small
    \renewcommand{\arraystretch}{1}
    \caption{Ablation Study on different solvers. Quantitative comparison between UniDB and UniDB++ with different orders.}
    \label{table_ablation_high_order}
    \vspace{-3mm}
    \fontsize{12pt}{14pt}\selectfont
    \resizebox{0.98\textwidth}{!}{
        \begin{tabular}{cccccccccccccc}
            \toprule[1.5pt]
            & & \multicolumn{4}{c}{\textbf{Image 4$\times$Super-Resolution}} & \multicolumn{4}{c}{\textbf{Raindrop Image Restoration}} & \multicolumn{4}{c}{\textbf{Image Deraining}} \\
            \cmidrule(lr){3-6} \cmidrule(lr){7-10} \cmidrule(lr){11-14}
            \textbf{Method} & \textbf{NFE} & \textbf{PSNR$\uparrow$} & \textbf{SSIM$\uparrow$} & \textbf{LPIPS$\downarrow$} & \textbf{FID$\downarrow$} & \textbf{PSNR$\uparrow$} & \textbf{SSIM$\uparrow$} & \textbf{LPIPS$\downarrow$} & \textbf{FID$\downarrow$} & \textbf{PSNR$\uparrow$} & \textbf{SSIM$\uparrow$} & \textbf{LPIPS$\downarrow$} & \textbf{FID$\downarrow$} \\
            \midrule[1pt]
            UniDB & 100 & 25.46 & 0.6856 & 0.179 & 16.21 & 28.58 & 0.8166 & 0.069 & 25.08 & 31.96 & 0.9028 & 0.046 & 18.14 \\
            \midrule
            UniDB++ & \multirow{2}{*}{10}  & \textbf{27.98} & \textbf{0.7940} & \textbf{0.208} & \textbf{15.78} & \textbf{30.66} & \textbf{0.9027} & \textbf{0.067} & \textbf{27.05} & \textbf{33.08} & \textbf{0.9288} & \textbf{0.047} & \textbf{17.55} \\
            UniDB++ (2nd) &  & 27.44 & 0.7808 & 0.210 & 16.09 & 29.74 & 0.8831 & 0.070 & 28.28 & 33.37 & 0.9341 & 0.050 & 18.77 \\
            \midrule
            UniDB++ & \multirow{2}{*}{20} & 27.38 & 0.7773 & \textbf{0.179} & 15.26 & 30.22 & 0.8915 & \textbf{0.065} & \textbf{24.95} & 32.57 & 0.9225 & \textbf{0.045} & \textbf{17.54} \\
            UniDB++ (2nd) &  & \textbf{27.64} & \textbf{0.7843} & 0.201 & \textbf{14.90} & \textbf{30.64} & \textbf{0.9022} & 0.068 & 25.57 & \textbf{32.84} & \textbf{0.9263} & 0.046 & 18.12 \\
            \midrule
            UniDB++ & \multirow{2}{*}{50} & 26.61 & 0.7535 & 0.159 & 14.98 & \textbf{29.64} & \textbf{0.8754} & 0.062 & 24.37 & 32.17 & 0.9158 & \textbf{0.043} & \textbf{17.19} \\
            UniDB++ (2nd) &  & \textbf{26.94} & \textbf{0.7566} & \textbf{0.147} & \textbf{14.10} & 29.51 & 0.8568 & \textbf{0.056} & \textbf{22.67} & \textbf{32.26} & \textbf{0.9184} & 0.044 & 17.21 \\
            \midrule
            UniDB++& \multirow{2}{*}{100} & 26.23 & 0.7403 & \textbf{0.154} & 15.50 & 29.42 & 0.8659 & \textbf{0.061} & \textbf{25.42} & 32.00 & 0.9124 & \textbf{0.043} & 17.76 \\
            UniDB++ (2nd) &  & \textbf{26.63} & \textbf{0.7539} & 0.159 & \textbf{14.49}  & \textbf{29.71} & \textbf{0.8751} & 0.063 & 25.44 & \textbf{32.17} & \textbf{0.9156} & \textbf{0.043} & \textbf{16.74} \\
            \bottomrule[1.5pt]
        \end{tabular}}
\end{table*}

\subsection{SDE-Corrector Mechanism}\label{section_corrector}

Recent advances in fast sampling techniques and current fast samplers \cite{dpm-solver, dpm-solver-plus} usually exhibit notable performance degradation and struggle to generate high-quality samples in the regime of extremely few steps (e.g. 5 $\sim$ 8 steps) \cite{Unipc}. Inspired by UniPC \cite{Unipc}, we further adopt an SDE-Corrector for the reverse-time SDE \eqref{unidb_gou_reverse_sde} to improve the performance of UniDB++ on perceptual fidelity in fewer sampling steps.

Now, given any nonzero increasing sequence $r_1 < r_2 < \cdots < r_k = 1$, denote $\beta_{s_i} = \beta_{s} + r_i h$ where $h = \beta_{t} - \beta_{s}$, $s_i = t_{\beta}(\beta_{s_i})$ for $i \in [1, k]$, and $\mathbf{x}_{t}^{c}$ as the corrected state vector, the updating correction step of $k$-th order UniDB++ is formed as 
\begin{equation}
\begin{aligned}
\mathbf{x}_{t}^{c} &= \frac{\kappa_{t, \gamma} \kappa_{s} \rho_t}{\kappa_{s, \gamma} \kappa_{t} \rho_s} \mathbf{x}_{s}^{c} + \left( 1 - \frac{\kappa_{t, \gamma} \kappa_{s} \rho_t}{\kappa_{s, \gamma} \kappa_{t} \rho_s} + \frac{\kappa_{t, \gamma} \rho_t \kappa_{s}}{\kappa_{0, \gamma} \kappa_{t} \rho_s} - \frac{\kappa_{t, \gamma}}{\kappa_{0, \gamma}}\right) x_T \\
&+ \frac{\kappa_{t, \gamma}}{\kappa_{0, \gamma}} B(h_i) \sum_{i=1}^{k} \frac{c_i}{r_i} (\mathbf{x}_\theta(\mathbf{x}_{s_i}, x_T, s_i) - \mathbf{x}_\theta(\mathbf{x}_{s}, x_T, s)) \\
&+ \left(\frac{\kappa_{t, \gamma}}{\kappa_{0, \gamma}} - \frac{\kappa_{t, \gamma} \rho_t \kappa_{s}}{\kappa_{0, \gamma} \kappa_{t} \rho_s} \right) \mathbf{x}_\theta(\mathbf{x}_{s}, x_T, s) + \delta^{d}_{s:t, \gamma} \boldsymbol{z},
\end{aligned}
\end{equation}
where $B(h)$ can be any function of $h$ such that $B(h) = \mathcal{O}(h)$, $\boldsymbol{R}_k(h) = [(r_j h)^{i-1}]_{i,j} \in \mathbb{R}^{k \times k}$, $\boldsymbol{c}_k(h) = (c_1(h), \cdots, c_k(h))^{T}$, $\boldsymbol{g}_k(h) = (g_1(h), \cdots, g_k(h))^{T}$, $\boldsymbol{c}_k(h) = \boldsymbol{R}^{-1}_k(h) \boldsymbol{g}_k(h) / B(h)$, $g_n = h^{n} n! \phi_{n+1}(h)$, and $\phi_{n}(h)$ is defined as $\phi_{n+1}(h) = (1/n! - \phi_{n}(h)) / h$ with $\phi_{1}(h) = (1 - e^{-h}) / h$. Taking first-order UniDB++ with SDE-Corrector as an example, we provide the pseudo-code as Algorithm \ref{unidb_data_sde_solver_1_corrector_sampling}. Based on the exact solution \eqref{sde_solution} and taking full use of the previous model results $\mathbf{x}_\theta(\mathbf{x}_{s}, x_T, s)$, the SDE-Corrector could increase the precision order without adding additional model evaluations \cite{Unipc}.

\section{Experiments}

\subsection{Experiment Setup}
In this section, we evaluate our models in image restoration tasks including Image 4$\times$Super-resolution (DIV2K dataset \cite{DIV2K}), Image Deraining (Rain100H dataset \cite{Rain100H}), Image Inpainting (CelebA-HQ 256$\times$256 dataset \cite{CelebAHQ}), and Raindrop Image Restoration (Raindrop dataset \cite{Raindrop}). We take four evaluation metrics: Peak Signal-to-Noise Ratio (PSNR, higher is better) \cite{PSNR}, Structural Similarity Index (SSIM, higher is better) \cite{SSIM}, Learned Perceptual Image Patch Similarity (LPIPS, lower is better) \cite{LPIPS}, and Fréchet Inception Distance (FID, lower is better) \cite{FID}. Also, we report the Number of Function Evaluations (NFE, lower is better) as a computational efficiency metric as the speedup for the NFE is approximately the actual speedup of the inference time \cite{DBIM}. Particularly in Image Deraining task, we report PSNR and SSIM scores on the Y channel (YCbCr space) to be aligned with the baselines. As for the mask type in Inpainting task, we take 100 thin masks from \cite{Repaint} consistent with the baselines. Unless explicitly mentioned, all experiments in this section employ the UniDB-GOU with Euler sampling (simply denoted UniDB in the following sections) and the first-order UniDB++ without the SDE-Corrector mechanism.

\subsection{Main Results}
For comparison among SDE-based methods, we choose the training-based methods IR-SDE \cite{IR-SDE}, DDBMs \cite{DDBMs}, GOUB \cite{GOUB}, and the training-free methods MaRS (for accelerating IR-SDE sampling) \cite{MaRS} and DBIMs (for accelerating DDBMs sampling) \cite{DBIM} as baselines. The quantitative and qualitative results are illustrated in Table \ref{table_main_sde_all} and Figure \ref{fig_main_experiment}. To characterize the trade-off between computational efficiency and sample quality, we evaluate UniDB++ at different NFEs to systematically assess its performance in both efficiency for small NFEs and quality for large NFEs. As for the comparison among the training-based methods, It is observed that our model UniDB achieved state-of-the-art results in all indicators. Furthermore, our proposed training-free algorithm UniDB++ with only $\text{NFE}=5$ or $\text{NFE}=10$ already achieves similar results to these training-based baselines with $\text{NFE}=100$ and surpasses all the baselines with $\text{NFE}=20$ in all image restoration tasks. Meanwhile, UniDB++ with $\text{NFE}=50$ achieves better perceptual scores (LPIPS and FID). Quantitative results in Table \ref{table_main_sde_all} also reveals UniDB++’s performance varies by sampling regime: better pixel-level details (higher PSNR and SSIM) at low NFEs, and improved perceptual quality (higher LPIPS and FID) at large NFEs, indicating that UniDB++ can further preserve image details as the sampling steps increase. Both UniDB and UniDB++ also excel by delivering superior performance in both visual quality and detail such as noses, eyes, alphabets, and shrubs compared to other results.

\begin{figure*}[ht]
	\begin{minipage}{0.33\linewidth}
		\vspace{3pt}
		\centerline{\includegraphics[width=0.9\textwidth]{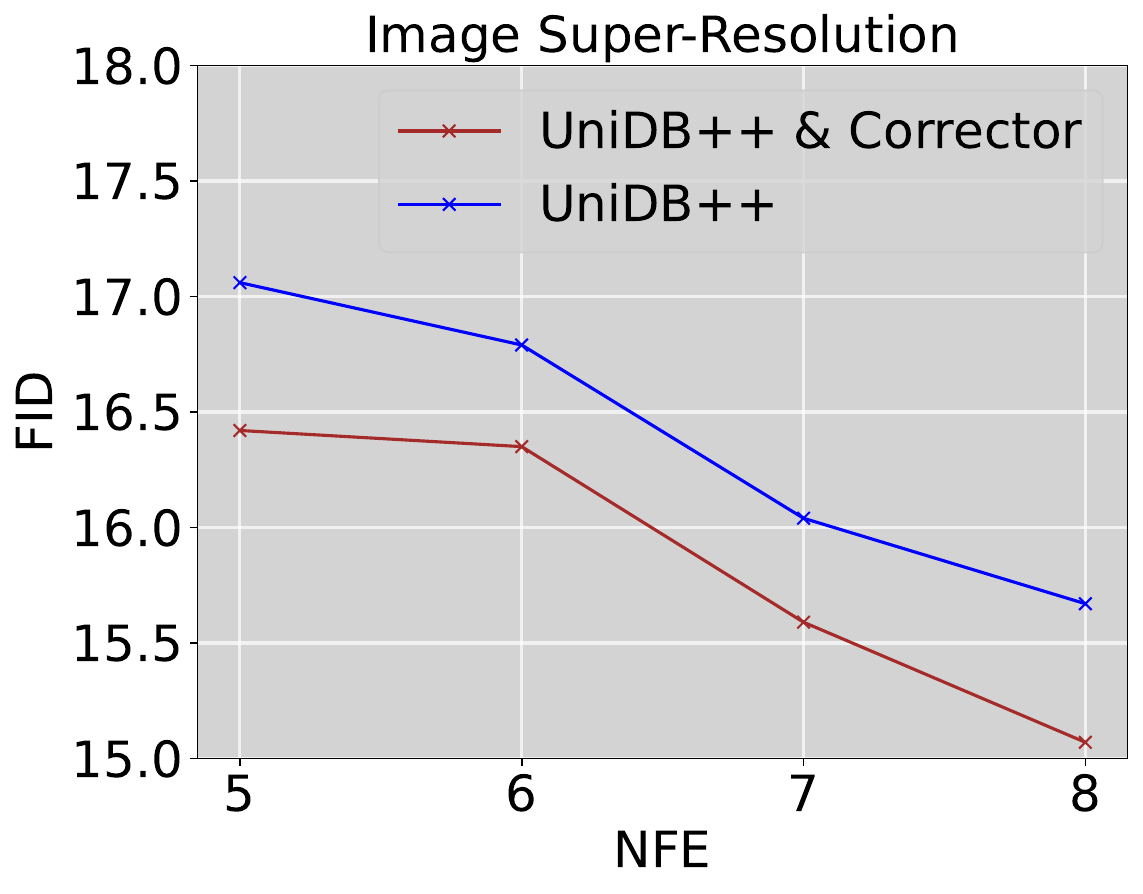}}
	\end{minipage}
    \begin{minipage}{0.33\linewidth}
		\vspace{3pt}
		\centerline{\includegraphics[width=0.9\textwidth]{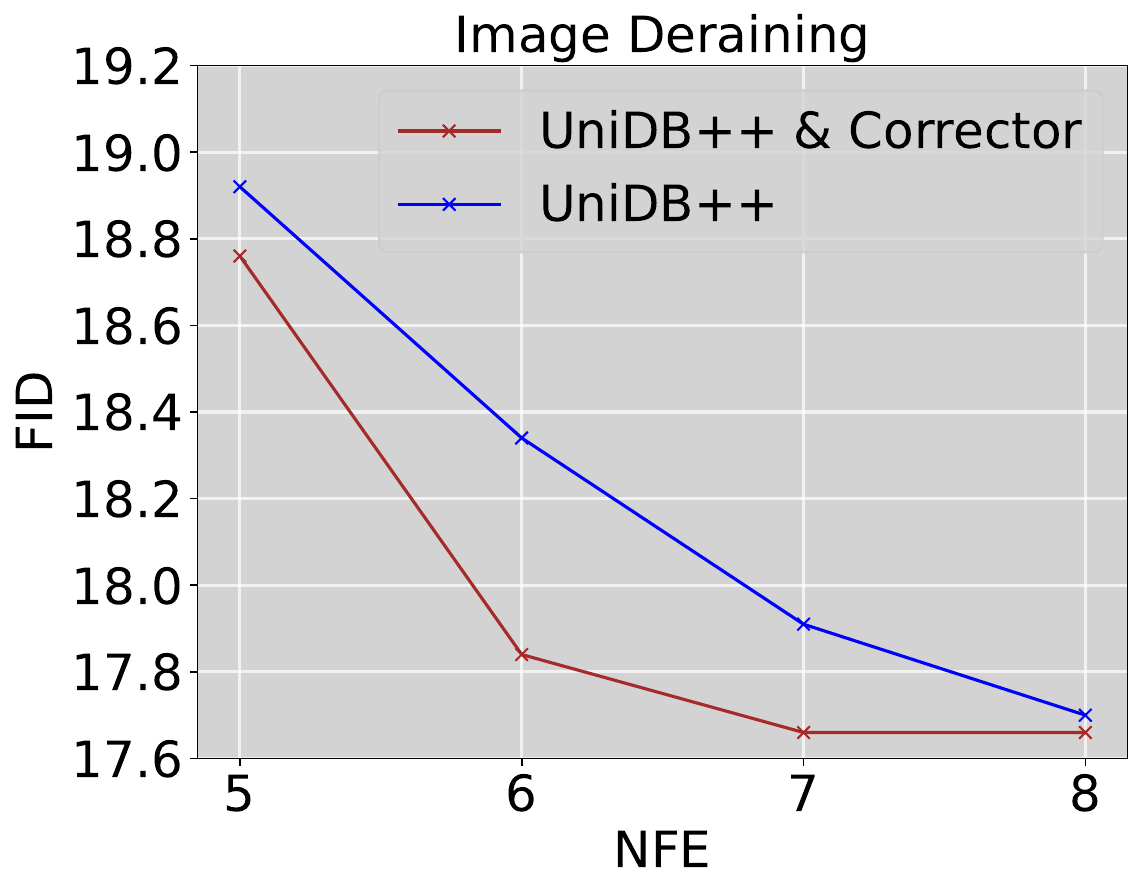}}
	\end{minipage}
	\begin{minipage}{0.33\linewidth}
		\vspace{3pt}
		\centerline{\includegraphics[width=0.9\textwidth]{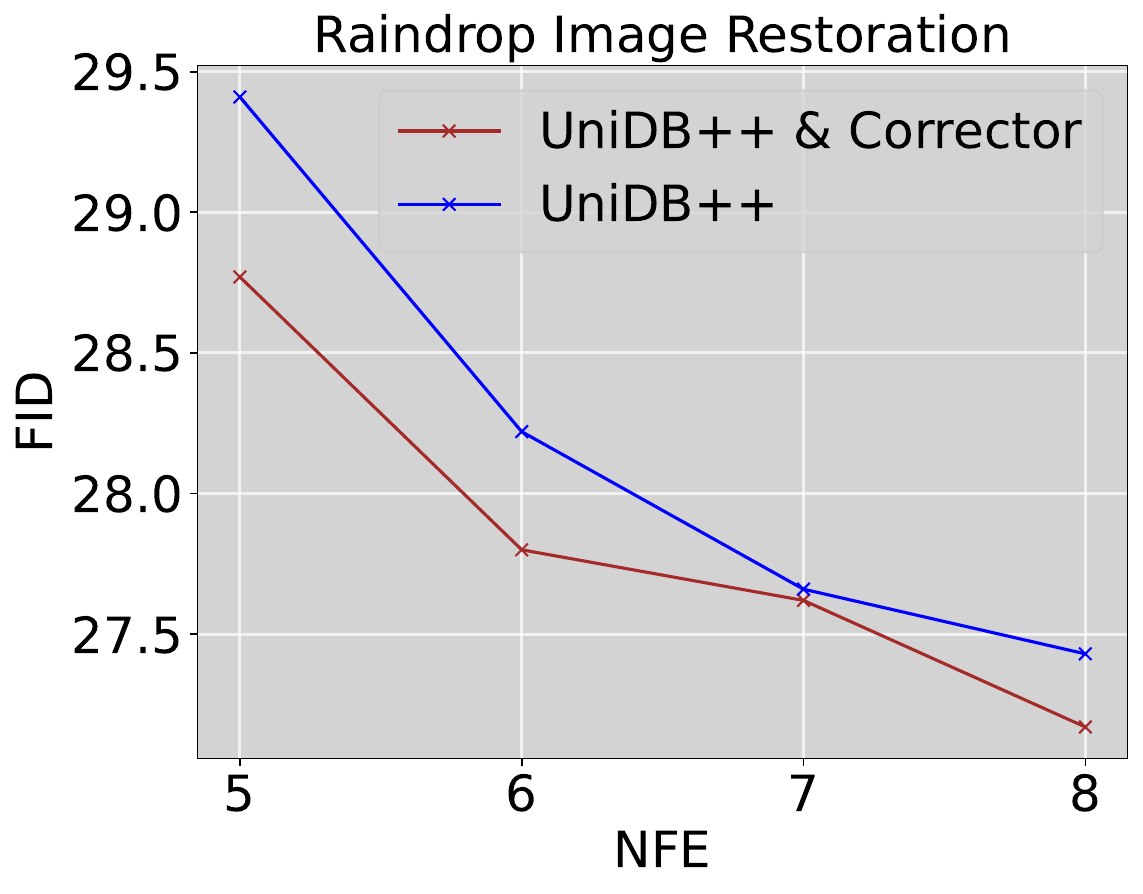}}
	\end{minipage}
    \vspace{-4mm}
	\caption{Ablation Study on SDE-Corrector. Qualitative comparison between UniDB++ without SDE-Corrector and UniDB++ with SDE-Corrector (both in first-order) (simply denoted as UniDB++ \& Corrector in the figures).}
	\label{fig_ablation_corrector_fid}
\end{figure*}

\begin{figure*}[h]
    \centering 
    \includegraphics[width=0.88\textwidth]{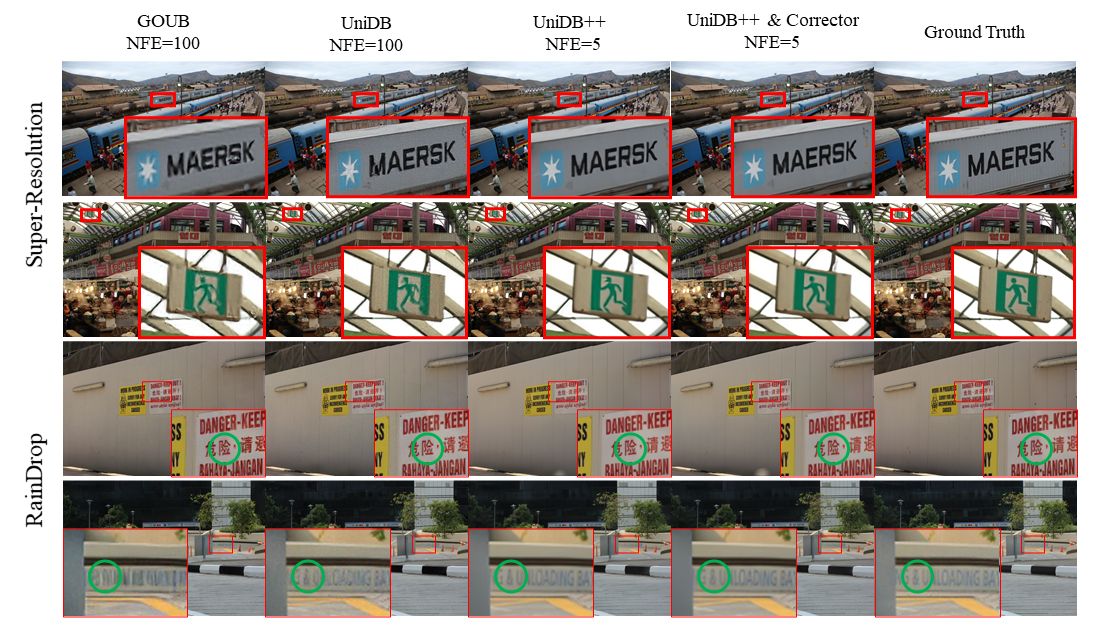}
    \vspace{-3mm}
    \caption{Qualitative comparison of visual results between GOUB, UniDB (ours), UniDB++ (ours) and UniDB++ with SDE-Corrector (simply denoted as UniDB++ \& Corrector in the figures) (ours) on Image Super-Resolution and Raindrop Image Restoration tasks.}
    \label{figure_ablation_corrector}
\end{figure*}

\subsection{Ablation Study}
\textbf{Effects on different Penalty Coefficient $\gamma$.} To evaluate the specific impact of different penalty coefficients $\gamma$ on model performance, we conducted the experiments with several different $\gamma$ and the final results are shown in Table \ref{table_ablation_gamma}. The results show that the choice of $\gamma$ significantly influences the model's performance on all tasks, different optimal $\gamma$ for different tasks, and our UniDB achieves the best performance in almost all metrics. Particularly in super-resolution tasks, we focus more on the significantly better perceptual scores (LPIPS and FID) \cite{IR-SDE}, demonstrating that UniDB ensures to capture more intricate image details and features as shown in Figure \ref{fig_main_experiment}. It is worth noting that when $\gamma$ is set to $1\times10^7$, our model outperforms the baselines across multiple tasks. These findings underscore the importance of carefully tuning $\gamma$ to achieve the best performance for specific tasks.

\textbf{Effects on different order solvers.} To systematically assess the performance of different numerical orders of UniDB++, we further conducted experiments on three image restoration tasks to evaluate the high-order (2nd) UniDB++ against the first-order case. For second-order UniDB++, we report the better results from the singlestep and multistep methods at the same NFEs for fair comparison on inference time. As demonstrated in Table \ref{table_ablation_high_order}, the second-order UniDB++ performs better with large NFEs ($\geq 50$), whereas the first-order UniDB++ delivers superior results with small NFEs ($\leq 20$). 

\textbf{Effects on SDE-Corrector.} We also conducted experiments on UniDB++ with or without the SDE-Corrector proposed in Section \ref{section_corrector}. As demonstrated in Figure \ref{figure_ablation_corrector}, UniDB++ equipped with SDE-Corrector exhibits superior perceptual quality compared to the standalone UniDB++ at low NFEs (5 $\leq$ NFE $\leq$ 8), especially in details like alphabets, signs and words. The corrected version of UniDB++ achieves lower FID scores in three image restoration tasks and better visual coherence in Figure \ref{figure_ablation_corrector}, indicating that the SDE-Corrector further enhances structural consistency and reduces artifacts.

\section{Conclusion}
In this paper, we present UniDB, a unified and fast-sampling diffusion bridge framework based on stochastic optimal control principles. Through this framework, we first unify and extend existing diffusion bridge models with Doob's $h$-transform by demonstrating that the diffusion bridge with Doob's $h$-transform can be viewed as a specific case when the terminal penalty coefficient approaches infinity. This insight helps elucidate why Doob's $h$-transform may lead to suboptimal image restoration, often resulting in blurred or distorted details. Furthermore, to avoid the computationally expensive inference in UniDB's Euler sampling methods, we propose UniDB++, an accelerated training-free sampling algorithm. The core contribution of UniDB++ lies in deriving exact closed-form solutions for UniDB’s reverse-time SDEs, which fundamentally addresses the error accumulation problem inherent to Euler approximations and enables high-fidelity generation with up to 20$\times$ fewer sampling steps. We replaced conventional noise prediction with a more stable data prediction model, along with an SDE-Corrector mechanism that maintains perceptual quality for extremely low-step regimes (5-8 steps). Comprehensive experimental evaluations confirm that our framework achieves state-of-the-art and superior results in both sample efficiency and image quality in various image restoration tasks. Future work will focus on improving the practicality of UniDB in different scientific fields, for example, in medical image restoration and imitation learning.
% Our experimental results underscore UniDB’s superiority and versatility across various image processing tasks, particularly in enhancing image details for more realistic outputs. Despite these advantages, UniDB, like other standard diffusion bridge models, faces the challenge of computationally intensive sampling processes, especially with high-resolution images or complex restoration tasks. Future work will focus on developing strategies to accelerate the sampling process, enhancing UniDB’s practicality, particularly for real-time applications. 

% \newpage
% \section*{Acknowledgments}

% {\appendix[Proof of the Zonklar Equations]
% Use $\backslash${\tt{appendix}} if you have a single appendix:
% Do not use $\backslash${\tt{section}} anymore after $\backslash${\tt{appendix}}, only $\backslash${\tt{section*}}.
% If you have multiple appendixes use $\backslash${\tt{appendices}} then use $\backslash${\tt{section}} to start each appendix.
% You must declare a $\backslash${\tt{section}} before using any $\backslash${\tt{subsection}} or using $\backslash${\tt{label}} ($\backslash${\tt{appendices}} by itself
%  starts a section numbered zero.)}

\bibliographystyle{IEEEtran}
\bibliography{reference}

% \newpage
{\appendices
% \section*{Proof of the First Zonklar Equation}
% \section*{Proof of the Second Zonklar Equation}
\section{Proof}

\subsection{Proof of Theorem \ref{theorem_4.1}} \label{proof_theorem_4.1}
\noindent \textbf{Theorem \ref{theorem_4.1}.} 
\textit{Consider the SOC problem \eqref{SOC_problem_generalized_ode}, denote $d_{t, \gamma} = \gamma^{-1} + e^{2\bar{f}_{T}} \bar{g}^2_{t:T}$, $\bar{f}_{s:t} = \int_{s}^{t} f_z dz$, $\bar{h}_{s:t} = \int_{s}^{t} e^{-\bar{f}_{z}} h_z dz$ and $\bar{g}^2_{s:t} = \int_{s}^{t} e^{-2\bar{f}_{z}}g^2_z dz$, denote $\bar{f}_{t}$, $\bar{h}_{t}$ and $\bar{g}^2_{t}$ for simplification when $s=0$, then the closed-form optimal controller $\mathbf{u}_{t,\gamma}^{*}$ is} 
\begin{equation}\tag{\ref{general_optimal_controller}}
\mathbf{u}_{t, \gamma}^{*} = g_t e^{\bar{f}_{t:T}} \frac{x_{T} - e^{\bar{f}_{t:T}} \mathbf{x}_t - \mathbf{m} e^{\bar{f}_{T}} \bar{h}_{t:T}}{d_{t, \gamma}},
\end{equation}
\textit{and the transition of $\mathbf{x}_t$ from $x_0$ and $x_T$ is}
\begin{equation}\tag{\ref{general_interpolant}}
\mathbf{x}_t = e^{\bar{f}_{t}} \Bigg(\frac{d_{t, \gamma}}{d_{0, \gamma}} x_0 + \frac{e^{\bar{f}_{T}} \bar{g}^2_{t}}{d_{0, \gamma}} x_T + \Big(\bar{h}_{t} - \frac{e^{2\bar{f}_{T}} \bar{h}_{T} \bar{g}^2_{t}}{d_{0, \gamma}}\Big) \mathbf{m}\Bigg). 
\end{equation}

\begin{proof}

According to Pontryagin Minimum Principle \cite{kirk2004optimal} recipe, one can construct the Hamiltonian: 
\begin{equation}
H(t,\mathbf{x}_t,\mathbf{u}_{t,\gamma},\mathbf{p}_t)=\frac{1}{2}\|\mathbf{u}_{t, \gamma}\|_{2}^{2}+ \mathbf{p}_t^{T} \left( f_t \mathbf{x}_t + h_t \mathbf{m} + g_t \mathbf{u}_t \right).
\end{equation}

By setting: 
\begin{equation}
\frac{\partial H}{\partial \mathbf{u}_{t, \gamma}} = 0 \quad \Rightarrow \quad \mathbf{u}_{t, \gamma}^{*} = - g_t \mathbf{p}_t.
\end{equation}

Then the value function becomes
\begin{equation}
\begin{aligned}
V^*=H(t,\mathbf{x}_t,\mathbf{p}_t,\mathbf{u}_{t, \gamma}^*) = -\frac{g^2_t}{2}\left\|\mathbf{p}_t\right\|^2_2 + f_t \mathbf{p}_t^{T} \mathbf{x}_t + h_t \mathbf{p}_t^{T} \mathbf{m}.
\end{aligned}
\end{equation}

Now, according to minimum principle theorem to obtain the following set of differential equations: 
\begin{equation}\label{mpt1_general}
\frac{\mathrm{d}\mathbf{x}_{t}}{\mathrm{d}t}=\nabla_{\mathbf{p}_t}H\left(\mathbf{x}_{t},\mathbf{p}_{t},\mathbf{u}_{t, \gamma}^{*},t\right)= - g^2_t \mathbf{p}_{t} + f_t\mathbf{x}_t + h_t\mathbf{m},
\end{equation}
\begin{equation}\label{mpt2_general}
\frac{\mathrm{d}\mathbf{p}_{t}}{\mathrm{d}t}= -\nabla_{\mathbf{x}_t}H\left(\mathbf{x}_{t},\mathbf{p}_{t},\mathbf{u}_{t}^{*},t\right) = -\mathbf{p}_{t} f_t,
\end{equation}
\begin{equation}\label{mpt3_general}
\mathbf{x}_{0} = x_{0},
\end{equation}
\begin{equation}\label{mpt4_general}
\mathbf{p}_{T}=\gamma \left(\mathbf{x}_T-x_{T}\right).
\end{equation}

Solving the Equation \eqref{mpt2_general}, we have:
\begin{equation}
\mathbf{p}_{t} = \mathbf{p}_{0} e^{-\bar{f}_{t}}, \ \mathbf{p}_{T} = \mathbf{p}_{0} e^{-\bar{f}_{T}}.
\end{equation}

Solve the Equation \eqref{mpt1_general}:
\begin{align*}
    &\frac{\mathrm{d} \mathbf{x}_t}{\mathrm{d} t} = f_t\mathbf{x}_t + h_t\mathbf{m} - g^2_t \mathbf{p}_{t} \\
    \Rightarrow \quad &\frac{\mathrm{d} (e^{-\bar{f}_{t}} \mathbf{x}_t)}{\mathrm{d} t} = e^{-\bar{f}_{t}} h_t \mathbf{m} - e^{-\bar{f}_{t}} g^2_t \mathbf{p}_{t}, \\
    \Rightarrow \quad &e^{-\bar{f}_{t}} \mathbf{x}_t - x_0 = \mathbf{m} \bar{h}_{t} - \mathbf{p}_{0} \bar{g}^2_{t}. \\
\end{align*}

Hence, we can get:
\begin{equation}\label{xt_general}
\mathbf{x}_t = e^{\bar{f}_{t}}x_0 + \mathbf{m} e^{\bar{f}_{t}} \bar{h}_{t} - \mathbf{p}_{T} e^{\bar{f}_{t}} e^{\bar{f}_{T}} \bar{g}^2_{t},
\end{equation}
\begin{equation}\label{x1_general}
\mathbf{x}_T = e^{\bar{f}_{T}}x_0 + \mathbf{m} e^{\bar{f}_{T}} \bar{h}_{T} - \mathbf{p}_{T} e^{2\bar{f}_{T}} \bar{g}^2_{T}.
\end{equation}

Take \eqref{x1_general} into \eqref{mpt4_general} and solve $\mathbf{p}_{T}$, 
\begin{align}\label{pT_general}
&\mathbf{p}_{T} = \gamma \left( e^{\bar{f}_{T}}x_0 + \mathbf{m} e^{\bar{f}_{T}} \bar{h}_{T} - \mathbf{p}_{T} e^{2\bar{f}_{T}} \bar{g}^2_{T} - x_{T} \right) \\
\Rightarrow \quad & \mathbf{p}_{T} = \frac{\gamma \left( e^{\bar{f}_{T}}x_0 + \mathbf{m} e^{\bar{f}_{T}} \bar{h}_{T} - x_{T} \right)}{1 + \gamma e^{2\bar{f}_{T}} \bar{g}^2_{T}}.
\end{align}

Also, take the Equation \eqref{pT_general} into the equation \eqref{xt_general}, 
\begin{equation}\label{36}
\begin{split}
    \mathbf{x}_t 
    &= e^{\bar{f}_{t}}x_0 + \mathbf{m} e^{\bar{f}_{t}} \bar{h}_{t} - e^{\bar{f}_{t}} e^{\bar{f}_{T}} \bar{g}^2_{t} \frac{e^{\bar{f}_{T}}x_0 + \mathbf{m} e^{\bar{f}_{T}} \bar{h}_{T} - x_{T}}{\gamma^{-1} + e^{2\bar{f}_{T}} \bar{g}^2_{T}} \\
    &= e^{\bar{f}_{t}} \Bigg(\frac{d_{t, \gamma}}{d_{0, \gamma}} x_0 + \frac{e^{\bar{f}_{T}} \bar{g}^2_{t}}{d_{0, \gamma}} x_T + \Big(\bar{h}_{t} - \frac{e^{2\bar{f}_{T}} \bar{h}_{T} \bar{g}^2_{t}}{d_{0, \gamma}}\Big) \mathbf{m}\Bigg). \\
\end{split}
\end{equation}

\begin{equation}
\begin{split}
\mathbf{u}^{*}_{t, \gamma} 
&= - g_t \mathbf{p}_{t} = - g_t e^{-\bar{f}_{t}} e^{\bar{f}_{T}} \frac{e^{\bar{f}_{T}}x_0 + \mathbf{m} e^{\bar{f}_{T}} \bar{h}_{T} - x_{T}}{\gamma^{-1} + e^{2\bar{f}_{T}} \bar{g}^2_{T}} \\
&= g_t e^{\bar{f}_{t:T}} \frac{x_{T} - e^{\bar{f}_{t:T}} \mathbf{x}_t - \mathbf{m} e^{\bar{f}_{T}} \bar{h}_{t:T}}{d_{t, \gamma}},
\end{split}
\end{equation}
with the fact \eqref{36}.
\end{proof}

\subsection{Proof of Theorem \ref{theorem_4.2}}\label{proof_theorem_4.2}
\noindent \textbf{Theorem \ref{theorem_4.2}.} 
\textit{For the SOC problem \eqref{SOC_problem_generalized_ode}, when $\gamma \to \infty$, the optimal controller becomes $\mathbf{u}^{*}_{t, \infty} = g_t \nabla_{\mathbf{x}_t} \log p(\mathbf{x}_T \mid \mathbf{x}_t)$, and the corresponding forward SDE is the same as Doob's $h$-transform as in \eqref{doob}.}

\begin{proof}
The optimal controller $\mathbf{u}^{*}_{t, \infty}$ is: 
\begin{equation}
\mathbf{u}^{*}_{t, \infty} = \lim_{\gamma \rightarrow \infty} \mathbf{u}^{*}_{t, \gamma} = g_t e^{\bar{f}_{t:T}} \frac{\mathbf{x}_{T} - e^{\bar{f}_{t:T}} \mathbf{x}_t - \mathbf{m} e^{\bar{f}_{T}} \bar{h}_{t:T}}{e^{2\bar{f}_{T}} \bar{g}^2_{t:T}}.
\end{equation}

As for $\mathbf{h}(\mathbf{x}_t, t, \mathbf{x}_T, T)$, consider $F(\mathbf{x}_t, t) = \mathbf{x}_t e^{-\bar{f}_t}$, according to the Ito differential formula, we get:
% TODO
\begin{equation}\label{general_transition}
\begin{aligned}
& \mathrm{d} F = -f_t \mathbf{x}_t e^{-\bar{f}_t} \mathrm{d} t + e^{-\bar{f}_t} \mathrm{d} \mathbf{x}_t\\
\Rightarrow \ & \mathrm{d} F = -f_t \mathbf{x}_t e^{-\bar{f}_t} \mathrm{d} t + e^{-\bar{f}_t} \Big( \left(f_t \mathbf{x}_t + h_t \mathbf{m}\right) \mathrm{d} t + g_t \mathrm{d} \boldsymbol{w}_t \Big), \\
\Rightarrow \ & \mathrm{d} F = h_t e^{-\bar{f}_t} \mathbf{m} \mathrm{d} t + e^{-\bar{f}_t} g_t \mathrm{d} \boldsymbol{w}_t, \\
\Rightarrow \ & \mathbf{x}_T e^{-\bar{f}_T} - \mathbf{x}_t e^{-\bar{f}_t} = \mathbf{m}\bar{h}_{t:T} + \int_{t}^{T} e^{-\bar{f}_z} g_z \mathrm{d} w_z, \\
\Rightarrow \ & \mathbf{x}_T \sim N\left( e^{\bar{f}_{t:T}} \mathbf{x}_t + \mathbf{m} e^{\bar{f}_T} \bar{h}_{t:T}, e^{2\bar{f}_T}\bar{g}^2_{t:T} \mathbf{I}\right),\\
\Rightarrow \ & \nabla_{\mathbf{x}_t} \log p(\mathbf{x}_T | \mathbf{x}_t) = \frac{e^{\bar{f}_{t:T}}\left(\mathbf{x}_T - e^{\bar{f}_{t:T}} \mathbf{x}_t - \mathbf{m} e^{\bar{f}_T} \bar{h}_{t:T}\right)}{e^{2\bar{f}_T}\bar{g}^2_{t:T}}, \\
\Rightarrow \ & \mathbf{u}^{*}_{t, \infty} = g_t \nabla_{\mathbf{x}_t} \log p(\mathbf{x}_T | \mathbf{x}_t) = g_t \mathbf{h}(\mathbf{x}_t, t, \mathbf{x}_T, T).
\end{aligned}
\end{equation}
% which concludes the proof of Theorem \ref{theorem_4.2}.
\end{proof}

\subsection{Proof of Proposition \ref{proposition_4.3}}\label{proof_proposition_4.3}
\textbf{Proposition \ref{proposition_4.3}} \textit{Consider the SOC problem \eqref{SOC_problem_generalized_ode}, denote $\mathcal{J}(\mathbf{u}_{t, \gamma}, \gamma) \triangleq \int_0^T \frac{1}{2} \left\|\mathbf{u}_{t, \gamma}\right\|_2^2 d t+\frac{\gamma}{2}\left\|\mathbf{x}_T^{u}-x_T\right\|_2^2$ as the overall cost of the system, $\mathbf{u}_{t, \gamma}^{*}$ as the optimal controller \eqref{general_optimal_controller}, then}
\begin{equation}
\mathcal{J}(\mathbf{u}_{t, \gamma}^{*}, \gamma) \le \mathcal{J}(\mathbf{u}_{t, \infty}^{*}, \infty). 
\end{equation}

\begin{proof}
According to \eqref{general_optimal_controller} and \eqref{general_interpolant}, simply denote $a = e^{\bar{f}_{T}} x_0 - x_T + \mathbf{m} e^{\bar{f}_{T}} \bar{h}_{T}$, 
\begin{align*}
& \mathbf{u}_{t, \gamma}^{*} = g_t e^{\bar{f}_{t:T}} \frac{x_{T} - e^{\bar{f}_{t:T}} \mathbf{x}_t - \mathbf{m} e^{\bar{f}_{T}} \bar{h}_{t:T}}{d_{t, \gamma}} \\
\Rightarrow \quad & \mathbf{u}_{t, \gamma}^{*} = -g_t e^{-\bar{f}_{t}} e^{\bar{f}_{T}} \frac{e^{\bar{f}_{T}} x_0 + \mathbf{m} e^{\bar{f}_{T}} \bar{h}_{T} - x_{T}}{d_{t, \gamma}}, \\
\Rightarrow \quad & \|\mathbf{u}_{t, \gamma}^{*}\|_2^2 = g^2_t e^{-2\bar{f}_{t}} e^{2\bar{f}_{T}} \frac{\|a\|_2^2}{d_{t, \gamma}^2} \\
\Rightarrow \quad & \|\mathbf{u}_{t, \infty}^{*}\|_2^2 = g^2_t e^{-2\bar{f}_{t}} e^{2\bar{f}_{T}} \frac{\|a\|_2^2}{(e^{2\bar{f}_{T}} \bar{g}^2_{T})^2}.
\end{align*}

Furthermore, take $t = T$ in \eqref{general_interpolant},
\begin{equation}
\begin{aligned}
& \quad \ \| \mathbf{x}_{T}^{u} - x_T \|_2^2 \\
&= \frac{\|e^{\bar{f}_{T}} x_0 - x_T + \mathbf{m} e^{\bar{f}_{T}} \bar{h}_{T}\|_2^2}{(1 + \gamma e^{2\bar{f}_{T}}\bar{g}^2_{T})^2} = \frac{\|a\|_2^2}{(1 + \gamma e^{2\bar{f}_{T}}\bar{g}^2_{T})^2 },
\end{aligned}
\end{equation}
\begin{equation}
\lim\limits_{\gamma \to \infty} \frac{\gamma}{2} \| \mathbf{x}_{T}^{u} - x_T \|_2^2 = \lim\limits_{\gamma \to \infty}\frac{\gamma}{2(1 + \gamma e^{2\bar{f}_{T}}\bar{g}^2_{T})^2} \|a\|_2^2 = 0.
\end{equation}

Hence, 
\begin{equation}
\begin{split}
& \quad \ \frac{1}{2}\int_{0}^{T} \left( \|\mathbf{u}_{t, \infty}^{*}\|_2^2 - \|\mathbf{u}_{t, \gamma}^{*}\|_2^2 \right) dt \\
&= \frac{1}{2}e^{2\bar{f}_{T}} \|a\|_2^2 \bar{g}^2_{T} \left( \frac{1}{(e^{2\bar{f}_{T}} \bar{g}^2_{T})^2} -\frac{1}{(\gamma^{-1}+ e^{2\bar{f}_{T}} \bar{g}^2_{T})^2} \right) \\ 
&= \frac{1}{2}\frac{1 + 2\gamma e^{2\bar{f}_{T}} \bar{g}^2_{T}}{(e^{2\bar{f}_{T}} \bar{g}^2_{T}) (1+ \gamma e^{2\bar{f}_{T}} \bar{g}^2_{T})^2} \|a\|_2^2 \\ 
&\ge \frac{1}{2}\frac{\gamma e^{2\bar{f}_{T}} \bar{g}^2_{T}}{(e^{2\bar{f}_{T}} \bar{g}^2_{T}) (1+ \gamma e^{2\bar{f}_{T}} \bar{g}^2_{T})^2} \|a\|_2^2 = \frac{\|a\|_2^2}{2}\frac{\gamma }{(1+ \gamma e^{2\bar{f}_{T}} \bar{g}^2_{T})^2} \\
&= \frac{\gamma}{2} \| \mathbf{x}_{T}^{u} - x_T \|_2^2 = \frac{\gamma}{2} \| \mathbf{x}_{T}^{u} - x_T \|_2^2 - \lim\limits_{\gamma \to \infty} \frac{\gamma}{2} \| \mathbf{x}_{T}^{u} - x_T \|_2^2,
\end{split}
\end{equation}
which implies
\begin{equation}
\begin{aligned}
& \frac{1}{2}\int_{0}^{T} \|\mathbf{u}_{t, \gamma}^{*}\|_2^2 dt + \frac{\gamma}{2} \| \mathbf{x}_{T}^{u} - x_T \|_2^2 \\
& \quad \quad \quad \quad \le \frac{1}{2}\int_{0}^{T} \|\mathbf{u}_{t, \infty}^{*}\|_2^2 dt + \lim\limits_{\gamma \to \infty} \frac{\gamma}{2} \| \mathbf{x}_{T}^{u} - x_T \|_2^2, \\
\Leftrightarrow \ & \mathcal{J}(\mathbf{u}_{t, \gamma}^{*}, \gamma) \le \mathcal{J}(\mathbf{u}_{t, \infty}^{*}, \infty).
\end{aligned}
\end{equation}
\end{proof}

\subsection{Proof of Theorem \ref{theorem_exact_solution}: Derivation of UniDB++ with Data Prediction Model}\label{appendix_data_solver}

\noindent \textbf{Theorem \ref{theorem_exact_solution}.} 
\textit{Given an initial value $\mathbf{x}_s$ at time $s \in [0, T]$, the exact solution $\mathbf{x}_t$ at time $t \in [0, s]$ of the reverse-time SDE \eqref{unidb_gou_reverse_sde_data} is:}
\begin{equation}\tag{\ref{sde_solution}}
\begin{gathered}
\begin{aligned}
\mathbf{x}_t = \frac{\kappa_{t, \gamma} \kappa_{s} \rho_t}{\kappa_{s, \gamma} \kappa_{t} \rho_s}&\mathbf{x}_s + \left( 1 - \frac{\kappa_{t, \gamma} \kappa_{s} \rho_t}{\kappa_{s, \gamma} \kappa_{t} \rho_s} + \frac{\kappa_{t, \gamma} \rho_t \kappa_{s}}{\kappa_{0, \gamma} \kappa_{t} \rho_s} - \frac{\kappa_{t, \gamma}}{\kappa_{0, \gamma}}\right)\mathbf{x}_T \\
&+ \frac{\kappa_{t, \gamma} \rho_t}{\kappa_{0, \gamma} \kappa_{t}} \int_{\beta_{s}}^{\beta_{t}} e^{\beta} \hat{\mathbf{x}}_\theta(\mathbf{x}_\beta, \mathbf{x}_T, \beta) d\beta + \delta^{d}_{s:t, \gamma} \boldsymbol{z},    
\end{aligned} \\
\begin{aligned}
(\delta^{d}_{s:t, \gamma})^2 = &\frac{\lambda^2 \kappa^2_{t, \gamma}\rho^2_t}{\kappa^2_{t}} \Bigg[E \frac{e^{2\bar{\theta}_{s}} - e^{2\bar{\theta}_{t}}}{(e^{2\bar{\theta}_{t}}-1)(e^{2\bar{\theta}_{s}}-1)} \\
&- D\log \frac{\kappa_{s, \gamma}\rho_t}{\kappa_{t, \gamma}\rho_s} - F( \frac{e^{-\bar{\theta}_{T}-\bar{\theta}_{t}}}{\kappa_{t, \gamma}} - \frac{e^{-\bar{\theta}_{T}-\bar{\theta}_{s}}}{\kappa_{s, \gamma}})\Bigg],
\end{aligned}
\end{gathered}
\end{equation}
\textit{where $c_1 = (\gamma\lambda^2)^{-1}e^{2\bar{\theta}_{T}}$, $c_2 = e^{2\bar{\theta}_{T}} - 1$, $D = -2c_1c_2/(c_1 + c_2)^3$, $E=c_2^2/(c_1 + c_2)^2$, $F=c_1^2/(c_1 + c_2)^2$, and $\boldsymbol{z} \sim \mathcal{N} (0, I)$ is a standard Guassian noise.}

\begin{proof}
% Review the reverse-time SDE with data precition model, 
% \begin{equation}\tag{\ref{unidb_gou_reverse_sde_data}}
% \begin{gathered}
% \begin{aligned}
% \mathrm{d} \mathbf{x}_t = \Bigg[&\left(\theta_t + \frac{g_t^2(1-\xi_t)}{\bar{\sigma}_{t}^{\prime2}}\right)x_T-\left(\theta_t - \frac{g_t^2}{\bar{\sigma}_{t}^{\prime2}} \right)\mathbf{x}_t \\
% &+ g_t\mathbf{u}_{t, \gamma}^{*}- \frac{g_t^2\xi_t}{\bar{\sigma}_{t}^{\prime2}} \mathbf{x}_\theta(\mathbf{x}_t, x_T, t) \Bigg] \mathrm{d} t + g_t \mathrm{d} \tilde{\boldsymbol{w}}_t,    
% \end{aligned} \\
% \mathbf{u}_{t, \gamma}^{*} = \frac{g_t e^{-2\bar{\theta}_{t:T}}}{\gamma^{-1} + \bar{\sigma}^2_{t:T}} (x_T - \mathbf{x}_t).
% \end{gathered}
% \end{equation}

First we list some definitions: 
\begin{equation}
\begin{gathered}
k_{t, \gamma} = \theta_t + \frac{g_t^2 e^{-2 \bar{\theta}_{t: T}}}{\gamma^{-1} + \bar{\sigma}_{t: T}^2} = \theta_t + \frac{2 \theta_t e^{-2 \bar{\theta}_{t: T}}}{(\gamma \lambda^2)^{-1} + 1 - e^{-2 \bar{\theta}_{t: T}}}, \\
\bar{k}_{s:t, \gamma} = \int_s^t k_{z, \gamma} dz, \ \bar{k}_{t, \gamma} = \int_0^t k_{z, \gamma} dz, \ e^{\beta_{t}} = \frac{\kappa_{t}}{\rho_t}.
\end{gathered}
\end{equation}

The reverse-time SDE with data precition model \eqref{unidb_gou_reverse_sde_data} can be written into
\begin{equation}\label{reverse_sde_data_simply}
\begin{aligned}
\mathrm{d} \mathbf{x}_t=\Bigg[&\left(k_{t, \gamma} - \frac{g_t^2}{\bar{\sigma}_{t}^{\prime2}} + \frac{g_t^2}{\bar{\sigma}_{t}^{\prime2}}\xi_t\right)x_T-\left(k_{t, \gamma} - \frac{g_t^2}{\bar{\sigma}_{t}^{\prime2}} \right)\mathbf{x}_t \\
&- \frac{g_t^2}{\bar{\sigma}_{t}^{\prime2}}\xi_t \mathbf{x}_\theta(\mathbf{x}_t, x_T, t) \Bigg] \mathrm{d} t + g_t \mathrm{d} \boldsymbol{w}_t
\end{aligned}
\end{equation}

Then, we start calculations, 
\begin{equation}
\begin{aligned}
\bar{k}_{s:t, \gamma} &= \int_s^t \left[\theta_z + 2 \theta_z \frac{e^{-2 \bar{\theta}_{z: T}}}{(\gamma \lambda^2)^{-1} + 1 - e^{-2 \bar{\theta}_{z: T}}} \right] dz \\
&= \bar{\theta}_{s:t} + \Big( -\log \left( (\gamma \lambda^2)^{-1} + 1 - e^x\right) \Big)\Big|_{2\bar{\theta}_{T:s}}^{2\bar{\theta}_{T:t}} \\
&= \bar{\theta}_{s:t} + \log \frac{(\gamma \lambda^2)^{-1} + 1 - e^{-2 \bar{\theta}_{s: T}}}{(\gamma \lambda^2)^{-1} + 1 - e^{-2 \bar{\theta}_{t: T}}},
\end{aligned}
\end{equation}

\begin{equation}
\Rightarrow \ e^{\bar{k}_{s:t, \gamma}} = e^{\bar{\theta}_{s:t}} \frac{(\gamma \lambda^2)^{-1} +1 - e^{-2 \bar{\theta}_{s: T}}}{(\gamma \lambda^2)^{-1} + 1 - e^{-2 \bar{\theta}_{t: T}}} = \frac{\kappa_{s, \gamma}}{\kappa_{t, \gamma}}.
\end{equation}

Consider
\begin{equation}
\frac{\mathrm{d} \beta_{t}}{\mathrm{d} t} = -\frac{2\theta_t(1-e^{-2\bar{\theta}_{T}})}{(1-e^{-2\bar{\theta}_{t}})(1-e^{-2\bar{\theta}_{t:T}})} = -\frac{g_{t}^{2}}{\bar{\sigma}_{t}^{\prime2}}.
\end{equation}
\begin{equation}
\int_{s}^{t} k_{\tau, \gamma} - \frac{g_\tau^2}{\bar{\sigma}_{\tau}^{\prime2}} d\tau = \bar{k}_{s:t, \gamma} + \int_{s}^{t} \frac{d \beta_{\tau}}{d\tau} d\tau = \bar{k}_{s:t, \gamma} + \beta_{t} - \beta_{s},
\end{equation}
\begin{equation}
e^{\int_{s}^{t} k_{\tau, \gamma} - \frac{g_\tau^2}{\bar{\sigma}_{\tau}^{\prime2}} d\tau} = e^{\bar{k}_{s:t, \gamma} + \beta_{t} - \beta_{s}} = \frac{\kappa_{s, \gamma}}{\kappa_{t, \gamma}}\frac{\kappa_{t}}{\rho_t}\frac{\rho_s}{\kappa_{s}} = \frac{\kappa_{s, \gamma} \kappa_{t} \rho_s}{\kappa_{t, \gamma} \kappa_{s} \rho_t},
\end{equation}
\begin{equation}
\begin{aligned}
&\quad \ \int_{s}^{t} e^{\bar{k}_{t:\tau, \gamma} + \beta_{\tau} - \beta_{t}}\frac{g_\tau^2}{\bar{\sigma}_{\tau}^{\prime2}}\xi_\tau d\tau = -\int_{s}^{t} \frac{\kappa_{t, \gamma} \kappa_{\tau} \rho_t}{\kappa_{\tau, \gamma} \kappa_{t} \rho_\tau} \frac{d \beta_{\tau}}{d\tau}\frac{\kappa_{\tau, \gamma}}{\kappa_{0, \gamma}} d\tau \\
&= -\frac{\kappa_{t, \gamma} \rho_t}{\kappa_{0, \gamma} \kappa_{t}} \int_{\beta_{s}}^{\beta_{t}} e^{\beta} d \beta = \frac{\kappa_{t, \gamma} \rho_t}{\kappa_{0, \gamma} \kappa_{t}} \left(\frac{\kappa_{s}}{\rho_s} - \frac{\kappa_{t}}{\rho_t}\right).
\end{aligned}
\end{equation}

% Consider $\delta^{d}_{s:t, \gamma}$,
\begin{equation*}
\begin{aligned}
& \int_s^t \frac{\kappa^2_{\tau} g^2_\tau}{\kappa^2_{\tau, \gamma} \rho^2_{\tau}} d\tau = \int_s^t \frac{(1 - e^{-2\bar{\theta}_{\tau:T}})^2 2\lambda^2 \theta_\tau (1 - e^{-2\bar{\theta}_{\tau}})^{-2}}{\left( (\gamma\lambda^2)^{-1} + 1 - e^{-2\bar{\theta}_{\tau:T}} \right)^2 e^{2\bar{\theta}_{\tau}} } d\tau \\
&= \int_s^t \frac{(e^{2\bar{\theta}_{T}} - e^{2\bar{\theta}_{\tau}})^2 2\lambda^2 \theta_\tau e^{2\bar{\theta}_{\tau}}}{\left( (\gamma\lambda^2)^{-1}e^{2\bar{\theta}_{T}} + e^{2\bar{\theta}_{T}} - e^{2\bar{\theta}_{\tau}} \right)^2 (e^{2\bar{\theta}_{\tau}} - 1)^2 } d\tau \\
&= \lambda^2 \int_{e^{2\bar{\theta}_{s}}-1}^{e^{2\bar{\theta}_{t}}-1} \frac{(e^{2\bar{\theta}_{T}} - 1 - x)^2}{\left( (\gamma\lambda^2)^{-1}e^{2\bar{\theta}_{T}} + e^{2\bar{\theta}_{T}} - 1 - x\right)^2 x^2} dx \\
&= \lambda^2 \int_{e^{2\bar{\theta}_{s}}-1}^{e^{2\bar{\theta}_{t}}-1} \frac{D}{x} + \frac{E}{x^2} + \frac{D}{c_1 + c_2 - x} + \frac{F}{(c_1 + c_2 - x)^2}dx \\
&= \lambda^2 \Bigg[ D\log \frac{e^{2\bar{\theta}_{t}}-1}{e^{2\bar{\theta}_{s}}-1} - E \left( \frac{1}{e^{2\bar{\theta}_{t}}-1} - \frac{1}{e^{2\bar{\theta}_{s}}-1} \right) \\
& \quad \quad \quad - D\log \frac{c_1 + c_2 + 1 - e^{2\bar{\theta}_{t}}}{c_1 + c_2 + 1 - e^{2\bar{\theta}_{s}}} \\
& \quad \quad \quad + F \left( \frac{1}{c_1 + c_2 + 1 - e^{2\bar{\theta}_{t}}} - \frac{1}{c_1 + c_2 + 1 - e^{2\bar{\theta}_{s}}} \right)\Bigg] \\
&= \lambda^2 \Bigg[ D\log \frac{\kappa_{s, \gamma}\rho_t}{\kappa_{t, \gamma}\rho_s} - E \left( \frac{1}{e^{2\bar{\theta}_{t}}-1} - \frac{1}{e^{2\bar{\theta}_{s}}-1} \right) \\
& \quad \quad \quad \quad \ + F \left( \frac{e^{-\bar{\theta}_{T}-\bar{\theta}_{t}}}{\kappa_{t, \gamma}} - \frac{e^{-\bar{\theta}_{T}-\bar{\theta}_{s}}}{\kappa_{s, \gamma}} \right)\Bigg],
\end{aligned}
\end{equation*}
where $D = -\frac{2c_1c_2}{(c_1 + c_2)^3}$, $E=\frac{c_2^2}{(c_1 + c_2)^2}$, and $F=\frac{c_1^2}{(c_1 + c_2)^2}$ can be derived by the method of undetermined coefficients. Hence, 
\begin{align*}
& \mathrm{d} \mathbf{x}_t=\Bigg[\left(k_{t, \gamma} - \frac{g_t^2}{\bar{\sigma}_{t}^{\prime2}} + \frac{g_t^2}{\bar{\sigma}_{t}^{\prime2}}\xi_t\right)x_T-\left(k_{t, \gamma} - \frac{g_t^2}{\bar{\sigma}_{t}^{\prime2}} \right)\mathbf{x}_t \\
&\quad \quad \quad \quad \quad \quad \quad \quad - \frac{g_t^2}{\bar{\sigma}_{t}^{\prime2}}\xi_t \mathbf{x}_\theta(\mathbf{x}_t, x_T, t) \Bigg] \mathrm{d} t + g_t \mathrm{d} \boldsymbol{w}_t \\
\Rightarrow & \mathrm{d} (e^{\bar{k}_{t, \gamma} + \beta_{t}}\mathbf{x}_t)= \Big[e^{\bar{k}_{t, \gamma}+ \beta_{t}}(k_{t, \gamma}- \frac{g_t^2}{\bar{\sigma}_{t}^{\prime2}} + \frac{g_t^2}{\bar{\sigma}_{t}^{\prime2}}\xi_t)x_T \\
&\quad \quad - e^{\bar{k}_{t, \gamma}+ \beta_{t}}\frac{g_t^2}{\bar{\sigma}_{t}^{\prime2}}\xi_t \mathbf{x}_\theta(\mathbf{x}_t, x_T, t) \Big] \mathrm{d} t + e^{\bar{k}_{t, \gamma} + \beta_{t}} g_t \mathrm{d} \boldsymbol{w}_t \\
\Rightarrow & \mathbf{x}_t = \frac{\kappa_{t, \gamma} \kappa_{s} \rho_t}{\kappa_{s, \gamma} \kappa_{t} \rho_s} \mathbf{x}_s + \int_{s}^{t} e^{\bar{k}_{t:\tau, \gamma} + \beta_{\tau} - \beta_{t}} (k_{\tau, \gamma}- \frac{g_\tau^2}{\bar{\sigma}_{\tau}^{\prime2}}) d\tau x_T\\
&\quad + \int_{s}^{t} e^{\bar{k}_{t:\tau, \gamma} + \beta_{\tau} - \beta_{t}} \frac{g_\tau^2}{\bar{\sigma}_{\tau}^{\prime2}}\xi_\tau d\tau x_T + \int_{s}^{t} \frac{\kappa_{t, \gamma}\kappa_{\tau}\rho_t}{\kappa_{t}\kappa_{\tau, \gamma}\rho_\tau} g_\tau d \boldsymbol{w}_\tau \\
&\quad - \int_{s}^{t} e^{\bar{k}_{t:\tau, \gamma} + \beta_{\tau} - \beta_{t}} \frac{g_\tau^2}{\bar{\sigma}_{\tau}^{\prime2}}\xi_\tau \mathbf{x}_\theta(\mathbf{x}_\tau, x_T, \tau) d\tau \\
\Rightarrow & \mathbf{x}_t = \frac{\kappa_{t, \gamma} \kappa_{s} \rho_t}{\kappa_{s, \gamma} \kappa_{t} \rho_s}\mathbf{x}_s + ( 1 - \frac{\kappa_{t, \gamma} \kappa_{s} \rho_t}{\kappa_{s, \gamma} \kappa_{t} \rho_s}) x_T \\
&\quad + \frac{\kappa_{t, \gamma} \rho_t}{\kappa_{0, \gamma} \kappa_{t}} (\frac{\kappa_{s}}{\rho_s} - \frac{\kappa_{t}}{\rho_t})x_T + \frac{\kappa_{t, \gamma}\rho_t}{\kappa_{t}} \sqrt{-\int_{s}^{t} \frac{\kappa^2_{\tau} g^2_\tau}{\kappa^2_{\tau, \gamma} \rho^2_{\tau}} d\tau } \boldsymbol{z} \\
&\quad + \frac{\kappa_{t, \gamma} \rho_t}{\kappa_{0, \gamma} \kappa_{t}} \int_{\beta_{s}}^{\beta_{t}} e^{\beta} \hat{\mathbf{x}}_\theta(\mathbf{x}_\beta, x_T, \beta) d\beta \\
\Rightarrow & \mathbf{x}_t = \frac{\kappa_{t, \gamma} \kappa_{s} \rho_t}{\kappa_{s, \gamma} \kappa_{t} \rho_s}\mathbf{x}_s + ( 1 - \frac{\kappa_{t, \gamma} \kappa_{s} \rho_t}{\kappa_{s, \gamma} \kappa_{t} \rho_s} + \frac{\kappa_{t, \gamma} \rho_t \kappa_{s}}{\kappa_{0, \gamma} \kappa_{t} \rho_s} - \frac{\kappa_{t, \gamma}}{\kappa_{0, \gamma}})x_T \\
&\quad \quad \quad + \frac{\kappa_{t, \gamma} \rho_t}{\kappa_{0, \gamma} \kappa_{t}} \int_{\beta_{s}}^{\beta_{t}} e^{\beta} \hat{\mathbf{x}}_\theta(\mathbf{x}_\beta, x_T, \beta) d\beta + \delta^{d}_{s:t, \gamma} \boldsymbol{z},
\end{align*}
where $\boldsymbol{z} \sim \mathcal{N} (0, I)$.
\end{proof}

\subsection{Derivation of UniDB++'s updating rule}\label{appendix_updating_rule}

\textbf{UniDB++ (1st-order).} Denote $M+1$ time steps $\left\{t_i\right\}_{i=0}^M$ decreasing from $t_0=T$ to $t_M=0$, directly take $\hat{\mathbf{x}}_\theta(\mathbf{x}_\beta, \mathbf{x}_T, \beta) \approx \mathbf{x}_\theta(\mathbf{x}_s, \mathbf{x}_T, s)$ in \eqref{sde_solution}, then the updating rule is
\begin{equation}
\begin{aligned}
&\mathbf{x}_{t_{i}} = \frac{\kappa_{t_{i}, \gamma} \kappa_{t_{i-1}} \rho_{t_{i}}}{\kappa_{t_{i-1}, \gamma} \kappa_{t_{i}} \rho_{t_{i-1}}} \mathbf{x}_{t_{i-1}} \\
&+ \left( 1 - \frac{\kappa_{t_{i}, \gamma} \kappa_{t_{i-1}} \rho_{t_{i}}}{\kappa_{t_{i-1}, \gamma} \kappa_{t_{i}} \rho_{t_{i-1}}} + \frac{\kappa_{t_{i}, \gamma} \kappa_{t_{i-1}} \rho_{t_{i}}}{\kappa_{0, \gamma} \kappa_{t_{i}} \rho_{t_{i-1}}} - \frac{\kappa_{t_{i}, \gamma}}{\kappa_{0, \gamma}} \right) x_T \\
&+ \left(\frac{\kappa_{t_{i}, \gamma}}{\kappa_{0, \gamma}} - \frac{\kappa_{t_{i}, \gamma} \kappa_{t_{i-1}} \rho_{t_{i}}}{\kappa_{0, \gamma} \kappa_{t_{i}} \rho_{t_{i-1}}}\right) \mathbf{x}_\theta(\mathbf{x}_{t_{i-1}}, x_T, t_{i-1}) + \delta^{d}_{t_{i-1}:t_{i}, \gamma} \boldsymbol{z}
\end{aligned}
\end{equation}

\textbf{UniDB++ (2nd-order).} To form UniDB++ (2nd-order), we need to approximate $\hat{\mathbf{x}}^{(1)}_\theta(\mathbf{x}_\beta, x_T, \beta)$. 

\textbf{Single-Step.} Define $\bar{\beta} = \beta_{t_{i-1}} + r h_i$ where $r \in (0, 1)$, then: 
\begin{equation}
\begin{aligned}
\hat{\mathbf{x}}^{(1)}_\theta(\mathbf{x}_\beta, \mathbf{x}_T, \beta) &\approx \hat{\mathbf{x}}^{(1)}_\theta(\mathbf{x}_{\beta_{t_{i-1}}}, x_T, \beta_{t_{i-1}}) \\
&\approx \frac{\hat{\mathbf{x}}_{\theta}(\mathbf{x}_{\bar{\beta}}, x_T, \bar{\beta}) - \hat{\mathbf{x}}_{\theta}(\mathbf{x}_{\beta_{t_{i-1}}}, x_T, \beta_{t_{i-1}})}{\bar{\beta} - \beta_{t_{i-1}}} \\
&= \frac{\hat{\mathbf{x}}_{\theta}(\mathbf{x}_{\bar{\beta}}, x_T, \bar{\beta}) - \hat{\mathbf{x}}_{\theta}(\mathbf{x}_{\beta_{t_{i-1}}}, x_T, \beta_{t_{i-1}})}{rh_i},
\end{aligned}
\end{equation}

\begin{equation}
\int_{\beta_{t_{i-1}}}^{\beta_{t_{i}}} e^{\beta} (\beta - \beta_{t_{i-1}}) d\beta= e^{\beta_{t_{i}}} \left( e^{-h_i} + h_i - 1 \right).
\end{equation}

The updating rule can be
\begin{equation}
\begin{aligned}
&\mathbf{x}_{t_{i}} = \frac{\kappa_{t_{i}, \gamma} \kappa_{t_{i-1}} \rho_{t_{i}}}{\kappa_{t_{i-1}, \gamma} \kappa_{t_{i}} \rho_{t_{i-1}}}  \mathbf{x}_{t_{i-1}} \\
&+ \left( 1 - \frac{\kappa_{t_{i}, \gamma} \kappa_{t_{i-1}} \rho_{t_{i}}}{\kappa_{t_{i-1}, \gamma} \kappa_{t_{i}} \rho_{t_{i-1}}} + \frac{\kappa_{t_{i}, \gamma} \kappa_{t_{i-1}} \rho_{t_{i}}}{\kappa_{0, \gamma} \kappa_{t_{i}} \rho_{t_{i-1}}} - \frac{\kappa_{t_{i}, \gamma}}{\kappa_{0, \gamma}} \right) x_T  \\
&+ \left(\frac{\kappa_{t_{i}, \gamma}}{\kappa_{0, \gamma}} - \frac{\kappa_{t_{i}, \gamma} \kappa_{t_{i-1}} \rho_{t_{i}}}{\kappa_{0, \gamma} \kappa_{t_{i}} \rho_{t_{i-1}}}\right) \mathbf{x}_\theta(\mathbf{x}_{t_{i-1}}, x_T, t_{i-1}) \\
&+ \frac{\kappa_{t_{i}, \gamma}}{\kappa_{0, \gamma}} ( e^{-h_i} + h_i - 1 ) \frac{\hat{\mathbf{x}}_{\theta}(\mathbf{x}_{\bar{\beta}}, x_T, \bar{\beta}) - \mathbf{x}_\theta(\mathbf{x}_{t_{i-1}}, x_T, t_{i-1})}{r h_i} \\
&+ \delta^{d}_{t_{i-1}:t_{i}, \gamma} \boldsymbol{z},
\end{aligned}
\end{equation}

\textbf{Multi-Step.} We approximate $\hat{\mathbf{x}}^{(1)}_\theta(\mathbf{x}_\beta, x_T, \beta)$ as
\begin{equation}
\begin{aligned}
& \quad \ \hat{\mathbf{x}}^{(1)}_\theta(\mathbf{x}_\beta, x_T, \beta) \approx \hat{\mathbf{x}}^{(1)}_\theta(\mathbf{x}_{\beta_{t_{i-1}}}, x_T, \beta_{t_{i-1}}) \\
&\approx \frac{\hat{\mathbf{x}}_{\theta}(\mathbf{x}_{\beta_{t_{i-1}}}, x_T, \beta_{t_{i-1}}) - \hat{\mathbf{x}}_{\theta}(\mathbf{x}_{\beta_{t_{i-2}}}, x_T, \beta_{t_{i-2}})}{\beta_{t_{i-1}} - \beta_{t_{i-2}}} \\
&= \frac{\mathbf{x}_\theta(\mathbf{x}_{t_{i-1}}, x_T, t_{i-1}) - \mathbf{x}_\theta(\mathbf{x}_{t_{i-2}}, x_T, t_{i-2})}{h_{i-1}}. \\
\end{aligned}
\end{equation}

The updating rule can be
\begin{equation}
\begin{aligned}
&\mathbf{x}_{t_{i}} = \frac{\kappa_{t_{i}, \gamma} \kappa_{t_{i-1}} \rho_{t_{i}}}{\kappa_{t_{i-1}, \gamma} \kappa_{t_{i}} \rho_{t_{i-1}}} \mathbf{x}_{t_{i-1}} \\
&+ \left( 1 - \frac{\kappa_{t_{i}, \gamma} \kappa_{t_{i-1}} \rho_{t_{i}}}{\kappa_{t_{i-1}, \gamma} \kappa_{t_{i}} \rho_{t_{i-1}}} + \frac{\kappa_{t_{i}, \gamma} \kappa_{t_{i-1}} \rho_{t_{i}}}{\kappa_{0, \gamma} \kappa_{t_{i}} \rho_{t_{i-1}}} - \frac{\kappa_{t_{i}, \gamma}}{\kappa_{0, \gamma}} \right) x_T \\
&+ \left(\frac{\kappa_{t_{i}, \gamma}}{\kappa_{0, \gamma}} - \frac{\kappa_{t_{i}, \gamma} \kappa_{t_{i-1}} \rho_{t_{i}}}{\kappa_{0, \gamma} \kappa_{t_{i}} \rho_{t_{i-1}}}\right) \mathbf{x}_\theta(\mathbf{x}_{t_{i-1}}, x_T, t_{i-1}) \\
&+ \frac{\kappa_{t_{i}, \gamma}( e^{-h_i} + h_i - 1 )}{\kappa_{0, \gamma}h_{i-1}} \Delta_{i-1} + \delta^{d}_{t_{i-1}:t_{i}, \gamma} \boldsymbol{z}.
\end{aligned}
\end{equation}
where $\Delta_{i-1} = \mathbf{x}_\theta(\mathbf{x}_{t_{i-1}}, x_T, t_{i-1}) - \mathbf{x}_\theta(\mathbf{x}_{t_{i-2}}, x_T, t_{i-2})$.

\begin{algorithm}[ht]
   \caption{UniDB++ Sampling ($k=2$, single-step)}
   \label{unidb_data_sde_solver_2_single_step}
\begin{algorithmic}
    \STATE {\bfseries Input:} LQ images $\mathbf{x}_T$, data predicted model $\mathbf{x}_\theta (\mathbf{x}_{t}, x_T, t)$, a buffer $Q$, and $M+1$ time steps $\left\{t_i\right\}_{i=0}^M$ decreasing from $t_0=T$ to $t_M=0$.
    \FOR{$i=1$ {\bfseries to} $M$}
        \STATE Sample $\boldsymbol{z}_1, \boldsymbol{z}_2 \sim \mathcal{N} (0, I)$ if $i < M$, else $\boldsymbol{z}_1 = \boldsymbol{z}_2 = 0$.
        \STATE $\hat{\mathbf{x}}_{0|i-1} = \mathbf{x}_\theta(\mathbf{x}_{t_{i-1}}, x_T, t_{i-1})$.
        \STATE $s = t_{\beta} \left(\beta_{t_{i-1}} + r h_i\right)$.
        \STATE $\mathbf{y}_i = \frac{\kappa_{s, \gamma} \kappa_{t_{i-1}} \rho_{s}}{\kappa_{t_{i-1}, \gamma} \kappa_{s} \rho_{t_{i-1}}}  \mathbf{x}_{t_{i-1}} + ( 1 - \frac{\kappa_{s, \gamma} \kappa_{t_{i-1}} \rho_{s}}{\kappa_{t_{i-1}, \gamma} \kappa_{s} \rho_{t_{i-1}}} + \frac{\kappa_{s, \gamma} \kappa_{t_{i-1}} \rho_{s}}{\kappa_{0, \gamma} \kappa_{s} \rho_{t_{i-1}}} - \frac{\kappa_{s, \gamma}}{\kappa_{0, \gamma}} ) x_T + (\frac{\kappa_{s, \gamma}}{\kappa_{0, \gamma}} - \frac{\kappa_{s, \gamma} \kappa_{t_{i-1}} \rho_{s}}{\kappa_{0, \gamma} \kappa_{s} \rho_{t_{i-1}}}) \hat{\mathbf{x}}_{0|i-1} + \delta^{d}_{t_{i-1}:s, \gamma} \boldsymbol{z}_1$.
        \STATE Calculate $\mathbf{x}_{\theta}(\mathbf{y}_i, \mathbf{x}_T, s)$.
        \STATE $\mathbf{x}_{t_{i}} = \frac{\kappa_{t_{i}, \gamma} \kappa_{t_{i-1}} \rho_{t_{i}}}{\kappa_{t_{i-1}, \gamma} \kappa_{t_{i}} \rho_{t_{i-1}}} \mathbf{x}_{t_{i-1}} + ( 1 - \frac{\kappa_{t_{i}, \gamma} \kappa_{t_{i-1}} \rho_{t_{i}}}{\kappa_{t_{i-1}, \gamma} \kappa_{t_{i}} \rho_{t_{i-1}}} + \frac{\kappa_{t_{i}, \gamma} \kappa_{t_{i-1}} \rho_{t_{i}}}{\kappa_{0, \gamma} \kappa_{t_{i}} \rho_{t_{i-1}}} - \frac{\kappa_{t_{i}, \gamma}}{\kappa_{0, \gamma}} ) x_T  + (\frac{\kappa_{t_{i}, \gamma}}{\kappa_{0, \gamma}} - \frac{\kappa_{t_{i}, \gamma} \kappa_{t_{i-1}} \rho_{t_{i}}}{\kappa_{0, \gamma} \kappa_{t_{i}} \rho_{t_{i-1}}}) \hat{\mathbf{x}}_{0|i-1} + \frac{\kappa_{t_{i}, \gamma}}{\kappa_{0, \gamma}} ( e^{-h_i} + h_i - 1 ) \frac{\mathbf{x}_{\theta}(\mathbf{y}_i, \mathbf{x}_T, s) - \hat{\mathbf{x}}_{0|i-1}}{r h_i} + \delta^{d}_{t_{i-1}:t_{i}, \gamma} \boldsymbol{z}_2$.
   \ENDFOR
   \STATE \textbf{Return} HQ images $\tilde{\mathbf{x}}_0$.
\end{algorithmic}
\end{algorithm}

\begin{algorithm}[ht]
   \caption{UniDB++ Sampling ($k=2$, multi-step)}
   \label{unidb_data_sde_solver_2_multi_step}
\begin{algorithmic}
    \STATE {\bfseries Input:} LQ images $\mathbf{x}_T$, data predicted model $\mathbf{x}_\theta (\mathbf{x}_{t}, x_T, t)$, a buffer $Q$, and $M+1$ time steps $\left\{t_i\right\}_{i=0}^M$ decreasing from $t_0=T$ to $t_M=0$.
    \FOR{$i=1$ {\bfseries to} $M$}
        \STATE Sample $\boldsymbol{z} \sim \mathcal{N} (0, I)$ if $i < M$, else $\boldsymbol{z} = 0$.
        % \STATE Sample $\boldsymbol{z} \sim \mathcal{N} (0, I)$ if $i < M$, else $\boldsymbol{z} = 0$.
        \IF{i == 1}
            \STATE $\mathbf{x}_{t_{1}} = \frac{\kappa_{t_{1}, \gamma} \kappa_{t_{0}} \rho_{t_{1}}}{\kappa_{t_{0}, \gamma} \kappa_{t_{1}} \rho_{t_{0}}}\mathbf{x}_{t_{0}} + ( 1 - \frac{\kappa_{t_{1}, \gamma} \kappa_{t_{0}} \rho_{t_{1}}}{\kappa_{t_{0}, \gamma} \kappa_{t_{1}} \rho_{t_{0}}} + \frac{\kappa_{t_{1}, \gamma} \kappa_{t_{0}} \rho_{t_{1}}}{\kappa_{0, \gamma} \kappa_{t_{1}} \rho_{t_{0}}} - \frac{\kappa_{t_{1}, \gamma}}{\kappa_{0, \gamma}} ) x_T + \delta^{d}_{t_{0}:t_{1}, \gamma} \boldsymbol{z}$.
            \STATE $Q \leftarrow \mathbf{x}_\theta(\mathbf{x}_{t_{0}}, \mathbf{x}_T, t_{0})$.
        \ELSE
            \STATE Take $\hat{\mathbf{x}}_{0|i-2} = \mathbf{x}_\theta(\mathbf{x}_{t_{i-2}}, x_T, t_{i-2})$ from $Q$.
            \STATE $\hat{\mathbf{x}}_{0|i-1} = \mathbf{x}_\theta(\mathbf{x}_{t_{i-1}}, \mathbf{x}_T, t_{i-1})$ and $Q \leftarrow \hat{\mathbf{x}}_{0|i-1}$.
            \STATE $\mathbf{x}_{t_{i}} = \frac{\kappa_{t_{i}, \gamma} \kappa_{t_{i-1}} \rho_{t_{i}}}{\kappa_{t_{i-1}, \gamma} \kappa_{t_{i}} \rho_{t_{i-1}}} \mathbf{x}_{t_{i-1}} + ( 1 - \frac{\kappa_{t_{i}, \gamma} \kappa_{t_{i-1}} \rho_{t_{i}}}{\kappa_{t_{i-1}, \gamma} \kappa_{t_{i}} \rho_{t_{i-1}}} + \frac{\kappa_{t_{i}, \gamma} \kappa_{t_{i-1}} \rho_{t_{i}}}{\kappa_{0, \gamma} \kappa_{t_{i}} \rho_{t_{i-1}}} - \frac{\kappa_{t_{i}, \gamma}}{\kappa_{0, \gamma}} ) x_T  + (\frac{\kappa_{t_{i}, \gamma}}{\kappa_{0, \gamma}} - \frac{\kappa_{t_{i}, \gamma} \kappa_{t_{i-1}} \rho_{t_{i}}}{\kappa_{0, \gamma} \kappa_{t_{i}} \rho_{t_{i-1}}}) \hat{\mathbf{x}}_{0|i-1} + \frac{\kappa_{t_{i}, \gamma}}{\kappa_{0, \gamma}} ( e^{-h_i} + h_i - 1 ) \frac{\hat{\mathbf{x}}_{0|i-1} - \hat{\mathbf{x}}_{0|i-2}}{h_{i-1}} + \delta^{d}_{t_{i-1}:t_{i}} \boldsymbol{z}$.
        \ENDIF
   \ENDFOR
   \STATE \textbf{Return} HQ images $\tilde{\mathbf{x}}_0$.
\end{algorithmic}
\end{algorithm}

\section{Experimental and Implementation Details}\label{append_experiment_details}
In all image restoration tasks, we directly follow the framework of GOUB \cite{GOUB} and keep all the training parameters and pre-trained weighted checkpoints exactly the same as theirs. The same noise network \cite{DPS}, steady variance level $\lambda^2 = 30^2 / 255^2$, coefficient $e^{\bar{\theta}_T} = 0.005$, 128 patch size with 8 batch size when training, ADAM optimizer with $\beta_1 = 0.9$ and $\beta_2 = 0.99$ \cite{ADAM}, 1.2 million total training steps with $10^{-4}$ initial learning rate and decaying by half at 300, 500, 600, and 700 thousand iterations. We choose a flipped version of cosine noise schedule \cite{IDDPMs} for $\theta_t$, $g_t$ is determined according to the relationship $g_{t}^{2} = 2 \lambda^2 \theta_t$. As for time schedule, we directly take the naive uniform time schedule. All experiments are tested on a single NVIDIA H20 GPU with 96GB memory. 
}

\vfill

\end{document}